\documentclass[11pt]{article}

\usepackage[final]{acl}

\usepackage{times}
\usepackage{latexsym}
\usepackage{CJKutf8}
\usepackage{makecell}
\usepackage{fvextra}

\usepackage[most]{tcolorbox}

\usepackage[T1]{fontenc}

\usepackage[utf8]{inputenc}

\usepackage{microtype}

\usepackage{inconsolata}

\usepackage{graphicx}
\usepackage{booktabs}
\usepackage{stfloats}
\usepackage{multirow}
\usepackage{array}
\usepackage[table]{xcolor}
\usepackage{url}
\usepackage{amsmath}
\usepackage{xspace}
\usepackage{subcaption}
\usepackage{enumitem}
\usepackage{fontawesome5}

\newcommand{\name}{\textbf{\textsc{Dr.DocBench}}}

\newcommand{\zexue}[1]{}
\newcommand{\xinyan}[1]{}
\newcommand{\minglai}[1]{}

\definecolor{myblue}{RGB}{220,230,250}
\definecolor{myframe}{RGB}{70,90,140}

\title{\name{}: A Comprehensive \\ Benchmark for Expert-Level and Difficult Document Parsing}

\author{
Minglai Yang$^{1,8}$\thanks{Equal contribution.}\;
Xinyan Velocity Yu$^{5}$\footnotemark[1]\;
Pengyuan Li$^{7}$\;
Xinyu Guo$^{1,8}$\;
Zhenting Qi$^{6}$\;
Konwoo Kim$^{2}$\\
\textbf{Longtian Ye}$^{1,9}$\;
\textbf{Xiaolong Luo}$^{6}$\;
\textbf{Jinhe Bi}$^{11}$\;
\textbf{Henry Zhang}$^{10}$\;
\textbf{Haris Riaz}$^{8}$\;
\textbf{Xuan Zhang}$^{1}$\;
\textbf{Yunze Xiao}$^{1,4}$\\
\textbf{Bangya Liu}$^{1}$\;
\textbf{Tom Tang}$^{1}$\;
\textbf{Yunfei Zhao}$^{1}$\;
\textbf{Qunshu Lin}$^{1}$\;
\textbf{Zihan Wang}$^{1}$\;
\textbf{Minghao Liu}$^{1,\dagger}$\\
\textbf{Michael Lingzhi Li}$^{6}$\;
\textbf{Yilun Du}$^{6}$\;
\textbf{Jesse Thomason}$^{5}$\;
\textbf{Rogerio Feris}$^{7}$\;
\textbf{Alex Pentland}$^{3}$\;
\textbf{Zexue He}$^{2}$\thanks{Corresponding authors.}\\[0.3em]
$^{1}$2077AI\quad
$^{2}$Stanford University\quad
$^{3}$MIT\quad
$^{4}$Carnegie Mellon University\\
$^{5}$University of Southern California\quad
$^{6}$Harvard University\quad
$^{7}$IBM Research\\
$^{8}$University of Arizona\quad
$^{9}$Duke University\quad
$^{10}$UC Berkeley\quad
$^{11}$LMU Munich\\[0.3em]
\texttt{\{minglai, minghao\}@2077ai.com, xinyany@usc.com, zexueh@stanford.edu}\\[0.3em]
\faGlobe\,\,\url{https://www.2077ai.com/drdocbench/}\quad
\faGithub\,\,\url{https://github.com/2077AI/DrDocBench}
}

\begin{document}
\maketitle

\begin{abstract}
Document parsing and recognition are fundamental capabilities for vision-language models (VLMs) and document processing systems. However, existing Optical Character Recognition (OCR) and document parsing benchmarks are increasingly limited in coverage and difficulty: many focus on common document genres or uniformly sampled pages where modern parsers already perform strongly, while offering limited annotation for expert-domain structures such as chemical formulas, music notation, complex tables, and cross-page layouts. We introduce \name, a difficulty-aware benchmark for expert-level document parsing. Built from a large-scale multilingual book corpus, \name{} spans 52 BISAC subject domains and selects challenging documents through parser-failure-based sampling, targeting cases where multiple state-of-the-art systems struggle. It contains  4{,}514 annotated pages from long documents averaging around 100 pages, with 70k high-quality page- and block-level annotations for layout, reading order,  hierarchical relations, and domain-specific visual contents. Evaluations of pipeline-based parsers and general-purpose VLMs show that strong performance on existing benchmarks does not transfer to our expert-level document parsing. Our analysis reveals substantial failures across subjects, content types, and structural attributes, highlighting \name{} as a comprehensive testbed for diagnosing and advancing document intelligence.
\end{abstract}

\section{Introduction}
\label{sec:intro}

Document parsing is a prerequisite for the daily usage of vision language models in the real world. Robust document understanding requires recovering both content and structure from visually complex pages, enabling downstream tasks such as visual document retrieval~\citep{faysse2025colpali,choi-etal-2025-zero, zhang2026mldocragmultimodallongcontextdocument}, question answering~\citep{cho2024m3docragmultimodalretrievalneed, gong2025mhierragmultimodalragvisualrich, dong2025benchmarking}, and scientific reasoning~\citep{luo2026p1, chen2026scimdradvancingscientificmultimodal}. 

\begin{figure}[t]
  \centering
  \includegraphics[width=0.49\linewidth]{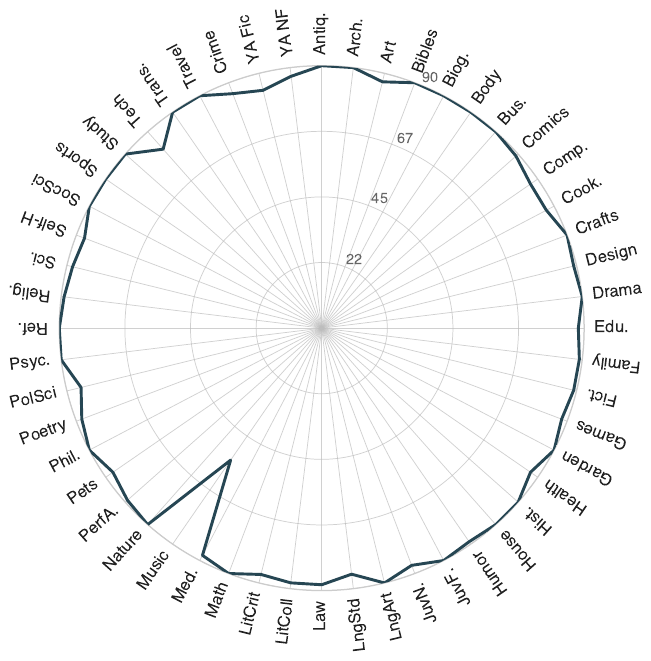}\hfill
  \includegraphics[width=0.49\linewidth]{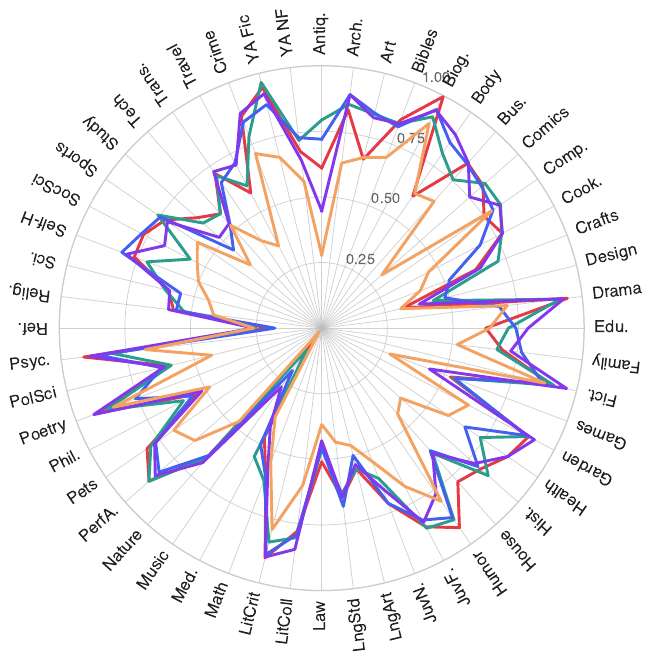}
  \caption{\textbf{Overview of \name{} across 52 BISAC subject domains.}
  \emph{Left}: annotated pages per subject (4{,}514 pages total).
  \emph{Right}: per-subject overall score for 5 frontier VLMs
  (\textcolor[HTML]{E63946}{\textbf{Claude Opus 4.6}},
  \textcolor[HTML]{2A9D8F}{\textbf{GPT-5.5}},
  \textcolor[HTML]{4361EE}{\textbf{Gemini 3.1 Pro}},
  \textcolor[HTML]{8338EC}{\textbf{Kimi}},
  \textcolor[HTML]{F4A261}{\textbf{GPT-4o}})
  across the 52 subjects, restricted to subjects with overall-metric coverage.}
  \label{fig:overview}
\end{figure}

\begin{figure*}
    \centering
    \includegraphics[width=1\linewidth]{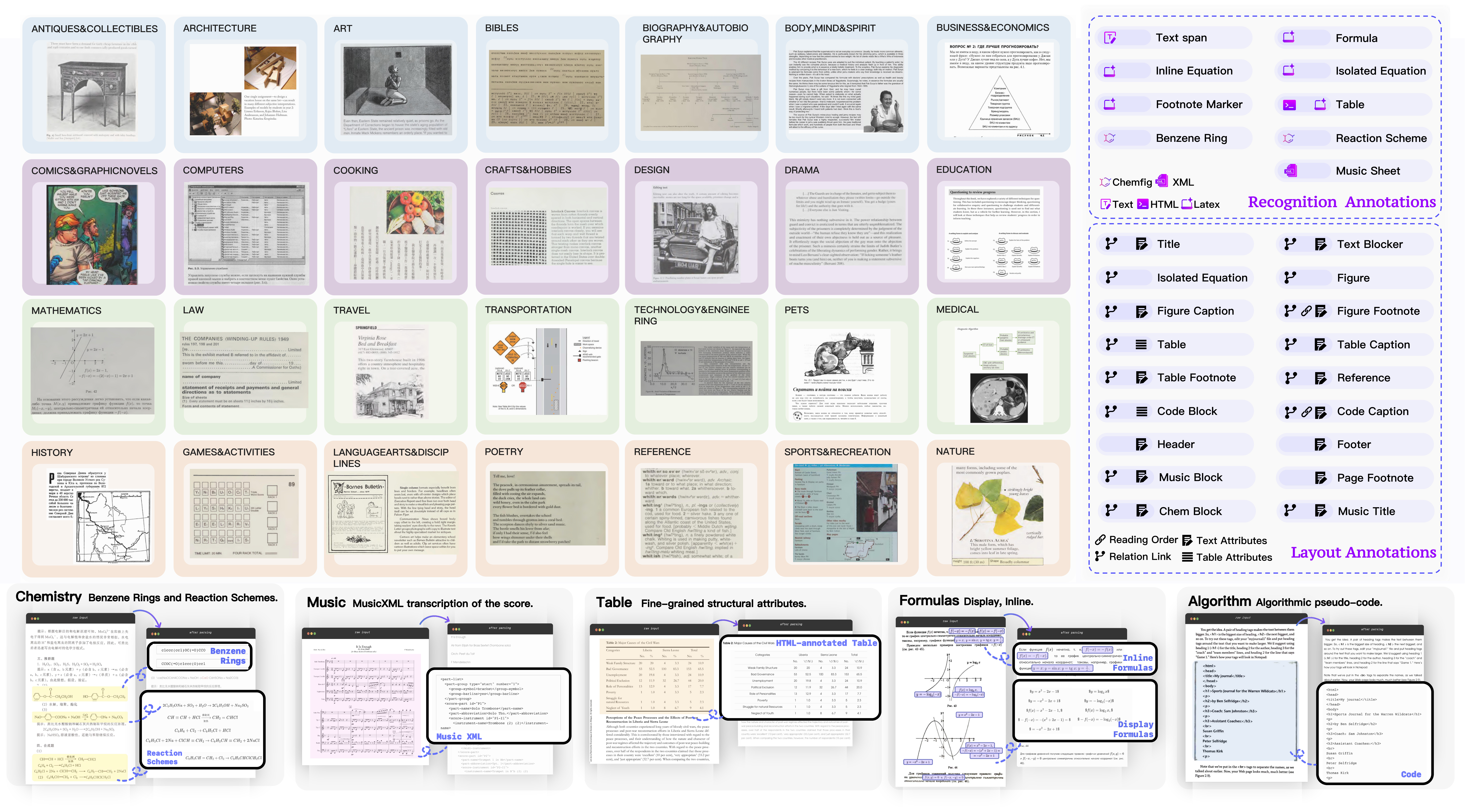}
\caption{
Overview of \name{}.
The benchmark spans diverse BISAC subject domains and provides fine-grained annotations for layout, recognition, and expert-domain structures, including chemistry diagrams, music notation, complex tables, formulas, and pseudo-code.
}
\label{fig:main}
\end{figure*}

While recent benchmarks~\citep{doclaynet2022, omnidocbench2024, fu2026ocrbench} have driven significant progress in evaluating document parsing for VLMs, they primarily focus on common genres. In familiar domains, models have already achieved ceiling performance~\citep{duan2026glmocrtechnicalreport, wei2026deepseek}. Consequently, current evaluations obscure important failures in the long tail of real-world documents. Furthermore, existing benchmarks often assess document understanding at a coarse, page-level granularity, lacking the detailed annotations required to penalize mistakes on complex, localized components (e.g., dense formulas, chemical structures, rotated tables, or domain-specific layouts). This creates an evaluation gap: although VLMs appear nearly saturated on standard tasks, their true capacity to parse expert-level, complex documents remains unmeasured.

Therefore, we introduce \name~(Figure~\ref{fig:main}), a subject-diverse,
difficulty-aware, expert-level benchmark designed to close this gap.
First, \name{} broadens data coverage to 52 BISAC subject domains with 4,514 annotated pages, spanning both common knowledge areas and long-tail expert domains. Second, it is constructed through difficulty-based data sourcing: candidate documents are first filtered using three state-of-the-art parsers, and challenging pages with high cross-parser disagreement are filtered for real human annotation. Third, we introduce expert-level annotations for domain-specific visual content. For example, chemistry pages may contain benzene rings and reaction schemes; music documents may contain visual scores that must be transcribed into structured music representations; and technical documents may contain dense formulas and complex tables. These contents cannot be captured by generic OCR labels alone, so we design annotation standards that translate expert visual languages into unified machine-understandable forms. Finally, the benchmark consists of ultra-long documents with an average length of $\sim$100 pages, enabling long-document parsing evaluation, such as reading order and cross-page continuity, that conventional single-page OCR benchmarks rarely address.

We conduct a systematic evaluation of both general-purpose and document-specialized VLMs on \name. We provide fine-grained analyses across subjects, content types, and structural attributes. Our results reveal that model strengths are highly uneven: some models perform relatively well on literature-style documents but struggle on technical and symbolic domains, while subjects such as \textsc{Reference} and \textsc{Music} remain difficult for nearly all systems. We further identify component-level failure modes, including color-dependent tables, dense formulas, and other visually complex structures. These findings show that strong performance on existing document benchmarks does not necessarily imply robust expert-level document parsing under our difficulty-aware setting. Overall, \name{} provides a comprehensive and fine-grained testbed for diagnosing the capabilities and limitations of current models and guiding future progress in document intelligence.\footnote{Data and code will be released upon acceptance.} %

\section{Related Work}
\label{sec:related}
\paragraph{Document parsing benchmarks.}
Several recent benchmarks have probed different facets of document recognition. OmniDocBench~\cite{omnidocbench2024} assembles pages from nine genres (academic papers, textbooks, exam sheets, slides, magazines, research reports, newspapers, handwritten notes, and colorful textbooks) to evaluate parsers across heterogeneous layouts. OCRBench~\cite{ocrbench2024}, together with its v2 successor~\cite{fu2026ocrbench}, expands the evaluation surface beyond plain text recognition to include table parsing, formula recognition, and reading-order prediction. Layout-centric efforts such as DocLayNet~\cite{doclaynet2022} and M$^6$Doc~\citep{Cheng_2023_CVPR} instead concentrate on region detection. MusiXQA~\citep{chen2025musixqaadvancingvisualmusic} evaluates MLLMs on music sheets via VQA over MusiXTeX~\citep{taupin1993musixtex} targets; we instead target end-to-end MusicXML~\citep{good2001musicxml, Hewlett2001TheVS} transcription. Our benchmark departs from prior work in two respects: rather than labeling pages with coarse genre buckets, we organize the corpus by \emph{document subject} using the BISAC heading system~\cite{bisac2025}, a 52-category taxonomy maintained for the North American book trade. After collapsing redundant codes, 52 BISAC top-level categories remain in our corpus, and we find that it is precisely the long tail of this distribution where current parsers degrade most sharply.

\paragraph{Document Parsing Systems.}
Document parsing systems have evolved from modular OCR pipelines to end-to-end VLM-based parsers. Traditional OCR systems~\citep{smith2007Tesseract, du2020ppocrpracticalultralightweight} decompose parsing into a multi-step pipeline of text detection, geometric correction, and text recognition. While efficient for standard OCR tasks, this approach is less flexible for interpreting complex document structures. Recent VLM-based specialized parsers~\citep{cui2026paddleocrvl15multitask09bvlm, niu2025mineru25decoupledvisionlanguagemodel, wang2026mineru25pro} are proposed to handle richer document structures, and therefore
are strong on targeted parsing tasks such as layout analysis, table extraction, and formula recognition, but are usually optimized around predefined parsing pipelines or output formats.
Concurrently, general-purpose VLMs such as the GPT family models~\citep{openai_2024_gpt4o, openai_2026_gpt55} and Qwen family models~\citep{bai2023qwenvlversatilevisionlanguagemodel, qwen3.5} are trained on large-scale multimodal corpora and further aligned for instruction following, where a variety of OCR-style data are often included in their training data mixtures. This not only enables instruction-guided, flexible OCR but also further unlocks advanced OCR-based multimodal reasoning, although their behavior can remain less predictable than specialized parsers.

\section{\name}
We design \name~to evaluate document parsing beyond common, increasingly saturated document genres and content types. 
\name~targets documents whose information is encoded through expert visual conventions, dense structure, and long-range layouts. We combine broad subject coverage, difficulty-aware sampling, and expert-level annotation to carefully curate such a benchmark.

\subsection{Difficulty-Aware Sampling}
\label{sec:filtering}

The source images of \name{} are from 
a large-scale book corpus, and further organised into using the BISAC taxonomy, spanning 52 subject domains across scientific, technical, artistic, educational, historical, literary, and reference materials.

Uniformly sampled pages are insufficient, since many common layouts are already handled well by current sota parsing systems. To focus annotation on genuinely difficult cases, \name{} uses a parser-failure-based sampling pipeline.

For each candidate book PDF, we run multiple strong document parsing systems independently, including PaddleOCR-VL-1.5~\citep{cui2026paddleocrvl15multitask09bvlm}, MinerU-VLM~\citep{wang2026mineru25pro}, and Gemini-3.1-pro-preview~\citep{gemini_team_2026_gemini31pro}, to get inter-parser disagreement score using normalized edit distance between parser outputs. Documents with low disagreement are discarded as these systems have converged on so less challenging, while high-disagreement documents are retained within each subject domain for next-step expert annotation. In practice, high-disagreement samples are often noticed to contain dense formulas, unusual layouts, multilingual content, complex tables, music notation, chemical diagrams, or visually crowded reference pages.

This filtering strategy makes \name{} difficulty-aware: the benchmark concentrates on documents where existing systems are unreliable, while maintaining balanced subject coverage. The parsers used for sampling are used only for data selection and are distinct
from the broader evaluation suite in Sec.~\ref{sec:experiments}, so
the filter neither advantages nor disadvantages any benchmarked models. Final evaluation is performed against human-verified ground truth.

\subsection{Annotation Schema}
\label{sec:schema}

\name{} provides\textit{ page-} and \textit{block-level annotations} that support both end-to-end evaluation and fine-grained error analysis. Each page is represented as a structured document object containing block regions, block types, transcribed content, reading order, and structural relations.

The schema covers common document elements such as paragraphs, titles, figures, tables, formulas, code blocks, algorithms, etc. It also captures relations that are critical for long-document parsing, including caption-to-object links, parent--child relations, and continuation links across columns or pages. These annotations allow evaluation to go beyond plain OCR accuracy and measure whether a system preserves document structure.

A central contribution of \name{} is its annotation of expert-domain content that prior benchmarks often ignore or collapse into generic categories. \textit{Chemical structures}, including benzene rings, fused-ring systems, and reaction schemes, are explicitly annotated rather than treated as ordinary figures. Particularly, \textit{Music sheets} are preserved using MusicXML~\citep{good2001musicxml, Hewlett2001TheVS}, while titles and credits are annotated as regular text. \textit{Mathematical} formulas are transcribed into normalized LaTeX with a fixed delimiter convention, ensuring that edit distance and structure-aware metrics remain comparable. For content-specific structures, tables are represented with HTML-style structure with cell spans, borders, and fine-grained layout attributes. Algorithms and pseudo-code are distinguished from regular code or text, and cross-block relations are used to connect each component that continues across columns or pages.

\subsection{Doctor Annotation and Quality Control}
\label{sec:quality_control}
Because many pages require domain knowledge (e.g., translate visual chemical formulas into text representations), \name{} uses a mixed annotation workflow. General document elements are annotated by general trained annotators, while specialized content is routed to domain experts with MS/PhD level education. For example, chemistry-majored annotators review molecular structures and reaction schemes, music-literate-majored annotators transcribe music scores, and LaTeX-fluent annotators handle dense mathematical content.

The annotation process begins with parser-generated sparse pre-labels, which are corrected and completed by human annotators. Annotators adjust bounding boxes, assign block categories, transcribe content, and add reading-order and structural relations. Expert-content blocks receive additional review by domain specialists.

We apply multiple quality-control checks before release. Most pages is double-annotated to identify ambiguous cases, and disagreements are resolved through adjudication. This process ensures that \name{} provides reliable ground truth not only for text recognition, but also for structural and domain-specific document understanding.

\subsection{Final Dataset}
\label{sec:statistics}

The final \name{} release contains \textbf{312 PDFs} and \textbf{4{,}514 annotated pages}, balanced at six PDFs per BISAC subject across \textbf{52 domains} and spanning \textbf{14 written languages} (predominantly English at $\sim$68\%). In total it carries over \text{70k fine-grained annotations}: $\sim$70k block-level labels across \text{21 block categories}. We use publicly available books and scanned documents solely for benchmark construction and research evaluation under fair-use principles, with attention to applicable access and licensing constraints.
Together with the expert-content tracks of Sec.~\ref{sec:schema}, these properties make \name{} a challenging benchmark for diagnosing the limitations of current VLMs and document parsing systems on the structures that prior benchmarks under-represent.

\section{Experiments}
\label{sec:experiments}

\subsection{Models}
We categorize document content extraction models into two main classes:
\paragraph{Specialized VLMs.} These are large multimodal models specifically trained for document parsing tasks. We experiment with PaddleOCR-VL-1.5~\citep{cui2026paddleocrvl15multitask09bvlm} and  MinerU~\citep{niu2025mineru25decoupledvisionlanguagemodel, wang2026mineru25pro}\footnote{\url{https://mineru.net/apiManage/docs}} through their official APIs, and they do not accept user prompts.

\paragraph{General VLMs.} This category comprises general-purpose large multimodal models inherently capable of document parsing. Our evaluation covers both open-source and proprietary models. The open-source selection includes Kimi-K2.5~\citep{kimiteam2026kimik25visualagentic}, Nemotron-nano-12b-v2-VL~\citep{nvidia2025nvidianemotronnanov2}, and three from the Qwen 3.5 family~\citep{qwen3.5}: Qwen-3.5-flash (a boosted 35B-A3B), Qwen-3.5-122B-A10B, and Qwen-3.5-plus (a boosted 397B-A17B). Additionally, we evaluate proprietary closed-source models: Gemini 3.1 Pro~\citep{gemini_team_2026_gemini31pro}, GPT-4o~\citep{openai_2024_gpt4o}, GPT-5.5~\citep{openai_2026_gpt55}, Claude Opus 4.6~\citep{anthropic_2026_claude_opus_4_6}, and Doubao-Seed 1.6-vision~\citep{bytedance_2025_doubao_seed_1_6_vision}.
Our inference prompt is in Appendix \ref{app:inference_prompt} (Figure \ref{fig:unified_prompt_part1},\ref{fig:unified_prompt_part2}). 

\subsection{Evaluation Metrics}

Following \citet{omnidocbench2024}, we employ different evaluation metrics tailored to different document elements:
\paragraph{Pure Text.} We calculate the Normalized Levenshtein Edit Distance~\citep{levenshtein1966binary}.
\paragraph{Tables.} We convert all tables to HTML format and report both Tree-Edit-Distance-based Similarity (TEDS) and TEDS\_S~\citep{zhong2020teds}.
\paragraph{Formulas.} We evaluate the formula using Character Detection Matching (CDM; \citet{wang2025cdm}) and Normalized Levenshtein Edit Distance. 

We report an \textit{Overall Score} (the higher the better) by mapping each component metric to a 0--100 accuracy scale: text and reading-order scores are $(1 - \text{edit distance}) \times 100$, formula scores are $\text{CDM} \times 100$, and table scores are $\text{TEDS} \times 100$. The overall score is the mean of the available component scores for each sample, with missing elements excluded from the denominator.

\subsection{Overall Results and Findings}
\begin{table*}[ht!]
  \centering
  \setlength{\tabcolsep}{3pt}
  \setlength{\aboverulesep}{1.5pt}
  \setlength{\belowrulesep}{1.5pt}
  \small
  \begin{tabular}{llcccccccc}
  \toprule
  \multirow{2}{*}{\textbf{Model}} &
  \multirow{2}{*}{\textbf{Size$^\dagger$}} &
  \multirow{2}{*}{\textbf{Access}} &
  \multicolumn{1}{c}{\textbf{Text}} &
  \multicolumn{2}{c}{\textbf{Formula}} &
  \multicolumn{2}{c}{\textbf{Table}} &
  \multicolumn{1}{c}{\textbf{Order}} &
  \multirow{2}{*}{\textbf{Overall}$\uparrow$} \\
  \cmidrule(lr){4-4} \cmidrule(lr){5-6} \cmidrule(lr){7-8} \cmidrule(lr){9-9}
  & & &
  Edit$\downarrow$ &
  Edit$\downarrow$ &
  CDM$\uparrow$ &
  TEDS$\uparrow$ &
  TEDS\textsubscript{S}$\uparrow$ &
  Edit$\downarrow$ & \\
  \midrule

  \multicolumn{10}{l}{\textit{Specialized VLMs}} \\
  \midrule
  MinerU 2.5              & 1.2B    & Open   & 0.33 & 0.51 & 24.15 & \textbf{55.85} & \textbf{63.70} & 0.30 & 54.37 \\
  PaddleOCR              & 0.9B    & Open   & \underline{0.73} & 0.42 & 30.73 & 51.79 & 59.94 & \underline{0.71} & 34.78 \\
  \midrule
  \multicolumn{10}{l}{\textit{General VLMs}} \\
  \midrule
  Qwen3.5-Flash            &  3B  & Open   & 0.26 & 0.32 & 35.69 & 46.24 & 54.38 & 0.26 & 57.51 \\
  Qwen3.5-122B-A10B        & 10B  & Open   & 0.23 & 0.32 & 32.54 & 38.31 & 44.99 & 0.23 & 56.32 \\
  Qwen3.5-Plus             & 17B    & Open   & 0.25 & \textbf{0.31} & 30.40 & 49.77 & 58.23 & 0.26 & 57.32 \\
  Nemotron-Nano-12B        & 12B  & Open   & 0.62 & \underline{0.76} & \underline{12.82} & \underline{27.03} & \underline{34.09} & 0.564 & \underline{30.33} \\
  Kimi-K2.5                & 32B   & Open   & 0.19 & 0.35 & 27.04 & 51.98 & 61.09 & 0.18 & 60.38 \\
  Claude Opus 4.6          & -    & Closed & 0.22 & 0.37 & 32.02 & 49.21 & 58.12 & 0.19 & 60.19 \\
  Doubao-Seed-1.6-Vision   & -    & Closed & 0.34 & 0.31 & \textbf{41.28} & 33.93 & 42.16 & 0.28 & 53.14 \\
  Gemini 3.1 Pro           & -    & Closed & 0.22 & 0.35 & 32.22 & 51.26 & 59.03 & 0.21 & 60.13 \\
  GPT-4o                   & -    & Closed & 0.37 & 0.47 & 36.02 & 30.14 & 38.62 & 0.31 & 49.73 \\
  GPT-5.5                  & -    & Closed & \textbf{0.19} & 0.36 & 34.55 & 48.90 & 58.96 & \textbf{0.17} & \textbf{61.94} \\
  \bottomrule
  \end{tabular}
  \caption{Overall Evaluation Results.
  $^\dagger$: activated Parameters. 
  \textbf{bold}: \textbf{best}, \underline{underlined}: \underline{worst}.}
  \label{tab:overall_unified_combined}
\end{table*}

Table~\ref{tab:overall_unified_combined} shows that no frontier VLM dominates expert-level document parsing. GPT-5.5, Kimi-K2.5, Claude Opus 4.6, and Gemini 3.1 Pro form a tight leading group, but their strengths vary by component: GPT-5.5 performs best on text extraction and reading order, while Kimi-K2.5 is more competitive on tables. Nemotron-Nano-12B, despite being a large pretrained VLM, is a clear low-performing outlier. Specialized parsers such as MinerU and PaddleOCR remain valuable for structured elements despite their smaller scale: MinerU achieves the best table scores, outperforming all general VLMs on both TEDS and TEDS\textsubscript{S}. However, because these systems are typically user-prompt-incapable, they perform worse on prompt-dependent metrics such as reading order and instruction-following text extraction.

\subsection{Per-Subject Breakdown and Analysis}
\begin{table}[t]
\centering
\setlength{\tabcolsep}{3pt}
\resizebox{\columnwidth}{!}{
\begin{tabular}{l c p{3.2cm} c p{3cm}}
\toprule
\multirow{2}{*}{\textbf{Model}} & \multicolumn{2}{c}{\textbf{Best 3}} & \multicolumn{2}{c}{\textbf{Worst 3}} \\
\cmidrule(lr){2-3} \cmidrule(lr){4-5}
 & \multicolumn{1}{c}{Score} & \multicolumn{1}{c}{Subject Names} & \multicolumn{1}{c}{Score} & \multicolumn{1}{c}{Subject Names} \\
\midrule
\textbf{Paddleocr} & 56.94 & YA Fic, Soc Sci, Hist & 13.79 & Drama, Ref, Fam \\
\textbf{Nemotron} & 59.15 & YA Fic, Edu, Biog & 14.41 & Antiq, Crafts, Ref \\
\textbf{Claude} & 96.13 & Biog, YA Fic, Drama & 35.07 & Ref, Des, Med \\
\textbf{Doubao} & 91.01 & YA Fic, Biog, Fic & 25.15 & Ref, Games, Antiq \\
\textbf{Gemini} & 90.84 & Biog, Arch, Lit Crit & 33.84 & Ref, Study, Law \\
\textbf{GPT-4o} & 86.83 & Biog, Fic, Poet & 27.16 & Ref, Games, Antiq \\
\textbf{GPT-5.5} & 93.08 & YA Fic, Drama, Fic & 38.24 & Ref, Des, Law \\
\textbf{Kimi} & 94.26 & Fic, Biog, Drama & 36.00 & Ref, Des, Med \\
\textbf{MinerU} & 91.77 & Drama, Fic, Biog & 23.13 & Comics, Ref, Law \\
\textbf{Qwen3.5-Flash} & 87.37 & Biog, YA Fic, Psych & 31.84 & Ref, Law, Games \\
\textbf{Qwen3.5-122B} & 92.35 & YA Fic, Humor, Fic & 29.58 & Study, Des, Games \\
\textbf{Qwen3.5-Plus} & 87.94 & Biog, YA Fic, Arch & 29.96 & Ref, Des, Law \\
\bottomrule
\end{tabular}
}
\caption{Model Best 3 and Worst 3 Subjects. Subject name abbreviations are listed in Appendix~\ref{app:abbrev}.}
\label{tab:best_worst_3_subjects_overall}
\end{table}

Table~\ref{tab:best_worst_3_subjects_overall} shows the best 3 subjects and the worst 3 subjects of VLMs on \name. Nearly all models are strongest on narrative or text-dense subjects such as \textsc{Biography \& Autobiography}, \textsc{Fiction}, and \textsc{Young Adult Fiction}, suggesting that current VLMs and parsers handle linear prose relatively well. 

In contrast, \textsc{Reference} is the most consistently difficult subject, appearing in the worst group for almost every model. \textsc{Design}, \textsc{Games}, \textsc{Medical}, and \textsc{Antiques \& Collectibles} recur as hard cases. These categories combine dense layouts, domain-specific notation, tables or lists that span multiple pages, or a visually irregular structure, indicating that expert-level document parsing remains limited less by ordinary text recognition than by structural and subject-specific understanding.

We provide the details of each models' worst five subjects in Appendix~\ref{app:detailed_worst_five}.

\subsection{Per-data-source Breakdown}

\begin{table*}[t]
  \centering
  \small
  \setlength{\tabcolsep}{3pt}
  \begin{tabular}{lrrrrrrrrrrr}
    \toprule
    Model & Type
      & All
      & Acad. Lit.
      & Book
      & Color. TB
      & Exam
      & Mag.
      & News.
      & Note
      & Res. Rep.
      & PPT \\
    \midrule
    MinerU            & \multirow{2}{*}{Spec.} & 0.325          & 0.227          & 0.321          & 0.347                    & 0.276          & 0.392          & 0.524          & 0.417          & 0.417                    & 0.245          \\
    PaddleOCR         &                        & 0.729          & 0.773          & 0.726          & 0.685                    & 0.744          & 0.707          & 0.817          & 0.744          & 0.543                    & 0.790          \\
    \midrule
    Qwen3.5-Flash     & \multirow{10}{*}{Gen.} & 0.258          & 0.209          & 0.261          & 0.184                    & 0.335          & 0.216          & 0.471          & 0.278          & 0.802                    & 0.163          \\
    Qwen3.5-122B-A10B &                        & 0.226          & 0.183          & 0.228          & 0.163                    & 0.277          & 0.192          & 0.370          & 0.275          & 0.904                    & 0.225          \\
    Qwen3.5-Plus      &                        & 0.249          & 0.223          & 0.251          & 0.162                    & 0.342          & 0.196          & 0.266          & 0.248          & \textbf{0.383}$^\dagger$ & 0.317          \\
    Nemotron          &                        & 0.622          & 0.577          & 0.625          & 0.546                    & 0.601          & 0.621          & 0.656          & 0.597          & 1.000                    & 0.394          \\
    Kimi              &                        & 0.193          & \textbf{0.174} & \textbf{0.190} & 0.147                    & 0.291          & 0.177          & 0.385          & 0.230          & 0.567                    & 0.162          \\
    Claude            &                        & 0.217          & 0.220          & 0.222          & \textbf{0.129}$^\dagger$ & 0.255          & \textbf{0.152} & \textbf{0.193} & 0.223          & 0.411                    & \textbf{0.101} \\
    Doubao            &                        & 0.344          & 0.332          & 0.346          & 0.258                    & 0.420          & 0.325          & 0.576          & 0.280          & 0.913                    & 0.475          \\
    Gemini            &                        & 0.219          & 0.214          & 0.216          & 0.160                    & 0.332          & 0.183          & 0.352          & 0.315          & 0.567                    & 0.223          \\
    GPT-4o            &                        & 0.366          & 0.311          & 0.372          & 0.249                    & 0.308          & 0.415          & 0.564          & 0.289          & 1.000                    & 0.503          \\
    GPT-5.5           &                        & \textbf{0.189} & 0.180          & 0.191          & \textbf{0.129}$^\dagger$ & \textbf{0.199} & 0.165          & 0.268          & \textbf{0.200} & \textbf{0.383}$^\dagger$ & 0.152          \\
    \bottomrule
  \end{tabular}
  \caption{%
    Text extraction edit distance by document data source ($\downarrow$ lower is better).
    \textbf{Bold}: best per column; $\dagger$ denotes a tie.
    \emph{Color.~TB}: colorful textbook;
    \emph{Acad.~Lit.}: academic literature;
    \emph{Res.~Rep.}: research report;
    \emph{PPT}: PPT2PDF.}
  \label{tab:text-source}
\end{table*}

We show the performance on different document source types in  Table~\ref{tab:text-source}. 
Research reports~(Res. Rep.) are the most challenging document type for text extraction across nearly all models. Doubao records the worst edit distance, while other frontier models, including GPT-4o (1.000), Gemini (0.567), and Kimi (0.567), also struggle on this document type. Only GPT-5.5 and Qwen3.5-plus maintain edit distances below 0.4.
In the meantime, presentation slides (\texttt{PPT2PDF}) show the largest inter-model performance variation among frontier VLMs: Claude achieves 0.101 while Doubao scores 0.475, corresponding to a 4.7$\times$ gap. 

\begin{table*}[t]
  \centering
  \resizebox{\textwidth}{!}{%
  \footnotesize
  \setlength{\tabcolsep}{4pt}
  \begin{tabular}{ll rr rrrr rr rr}
    \toprule
    & & \multicolumn{6}{c}{\textbf{Text}}
    & \multicolumn{4}{c}{\textbf{Table}} \\
    \cmidrule(lr){3-8}\cmidrule(lr){9-12}
    & & \multicolumn{2}{c}{\textit{RO Dist $\downarrow$}}
    & \multicolumn{4}{c}{\textit{Text Edit Dist $\downarrow$}}
    & \multicolumn{2}{c}{\textit{TEDS (\%) $\uparrow$}}
    & \multicolumn{2}{c}{\textit{TEDS (\%) $\uparrow$}} \\
    \cmidrule(lr){3-4}\cmidrule(lr){5-8}\cmidrule(lr){9-10}\cmidrule(lr){11-12}
    Model & Type
      & Single Col.
      & $\geq$2 Col.
      & Rot. Norm.
      & Rot. 90\textdegree
      & Rot. 270\textdegree
      & Rot. Horiz.
      & Full Line
      & Wireless
      & BG Yes
      & BG No \\
    \midrule
    MinerU            & \multirow{2}{*}{Spec.} & 0.237          & 0.357          & 0.354          & 0.539          & 0.563          & 0.546          & \textbf{54.0} & \textbf{78.8} & 32.3          & \textbf{67.9} \\
    PaddleOCR         &                        & 0.687          & 0.813          & 0.941          & 0.971          & 0.966          & 0.854          & 43.9          & 75.9          & 29.0          & 62.4          \\
    \midrule
    Qwen3.5-Flash     & \multirow{10}{*}{Gen.} & 0.197          & 0.357          & 0.408          & 0.518          & 0.576          & 0.478          & 39.8          & 55.1          & 26.3          & 49.1          \\
    Qwen3.5-122B-A10B &                        & 0.175          & 0.339          & 0.319          & 0.496          & 0.481          & 0.432          & 32.0          & 15.2          & 22.3          & 27.8          \\
    Qwen3.5-Plus      &                        & 0.222          & 0.329          & 0.416          & 0.502          & 0.527          & 0.370          & 49.1          & 77.0          & 29.5          & 63.7          \\
    Nemotron          &                        & 0.508          & 0.645          & 0.708          & 0.880          & 0.750          & 0.964          &  9.6          & 10.3          &  9.0          & 15.7          \\
    Kimi              &                        & \textbf{0.133} & 0.263          & 0.259          & 0.672          & 0.432          & 0.357          & 49.1          & 21.4          & 33.1          & 40.1          \\
    Claude            &                        & 0.158          & 0.256          & 0.257          & 0.269          & 0.293          & 0.413          & 43.0          & 46.9          & 26.4          & 49.3          \\
    Doubao            &                        & 0.213          & 0.384          & 0.477          & 0.761          & 0.757          & 0.545          & 29.0          &  8.1          & 16.7          & 22.6          \\
    Gemini            &                        & 0.177          & 0.269          & 0.319          & 0.446          & \textbf{0.277} & 0.341          & 41.6          & 77.1          & 29.2          & 60.8          \\
    GPT-4o            &                        & 0.233          & 0.420          & 0.479          & 0.814          & 0.758          & 0.763          & 17.1          & 11.2          & 12.8          & 18.7          \\
    GPT-5.5           &                        & 0.143          & \textbf{0.226} & \textbf{0.215} & \textbf{0.240} & 0.419          & \textbf{0.318} & 51.5          & 74.9          & \textbf{35.4} & 62.3          \\
    \bottomrule
  \end{tabular}%
  }
  \caption{%
    Per block type breakdown.
    For \textbf{Text}, we report \textit{Reading-Order (RO) edit distance} ($\downarrow$) by layout (Col.: columns) and \textit{text edit distance} ($\downarrow$) by block rotation (Norm.: normal, Horiz.: horizontal).
    For \textbf{Table}, we report TEDS (\%, $\uparrow$) by border type and by table background color.
    \textbf{Bold}: best per column; $\dagger$ denotes a tie.}
  \label{tab:combined-text-table}
\end{table*}

\subsection{Per-block Breakdown}
The block-level breakdown in Table \ref{tab:combined-text-table} shows that many errors come from layout and visual-structure ambiguity rather than plain text recognition alone. For reading order, single-column pages are consistently the easiest, while multi-column layouts substantially increase error for nearly all models. 
Similarly, rotated text is harder than normally oriented text, with 90-degree rotation often especially challenging, suggesting strong orientation biases in current models.

For tables, the clearest stressor is visual complexity and the presence of borders. Models also find tables with colored backgrounds harder to transcribe, as we observe a large and consistent drop in model performance, suggesting that background shading interferes with cell localization and content-structure alignment. Border style is more nuanced, implying that models rely on different cues for table reconstruction. In general, non-standard table layouts, especially list-style or vertical layouts, are more consistently challenging because they break the regular row-column assumptions used by most parsers.
\section{Additional Analysis}
\label{sec:analysis}

\subsection{Sensitivity to Multi-page Context Length}
\label{sec:context_sensitivity}
\begin{figure*}[tbhp!]
    \centering
    \includegraphics[width=\linewidth]{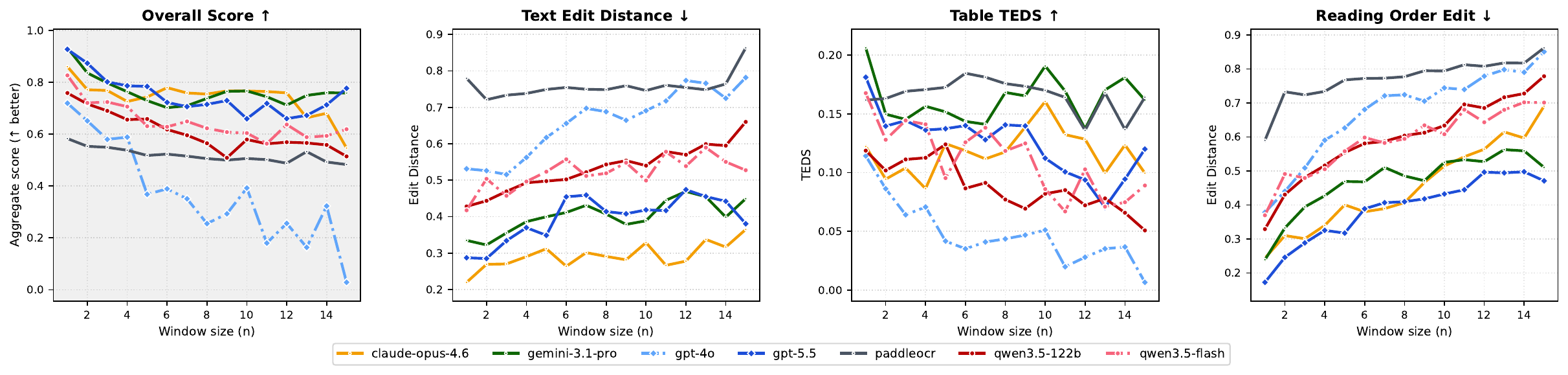}
    \caption{Impact of sliding window size ($n$ pages) on overall and per-metric performance across seven models. The top-left panel shows a normalized aggregate score; the remaining panels show individual metrics. Reading order degrades most consistently with increasing window size, while formula and table metrics are comparatively stable.}
    \label{fig:scaling_window}
\end{figure*}

\begin{figure}
    \centering
    \includegraphics[width=\linewidth]{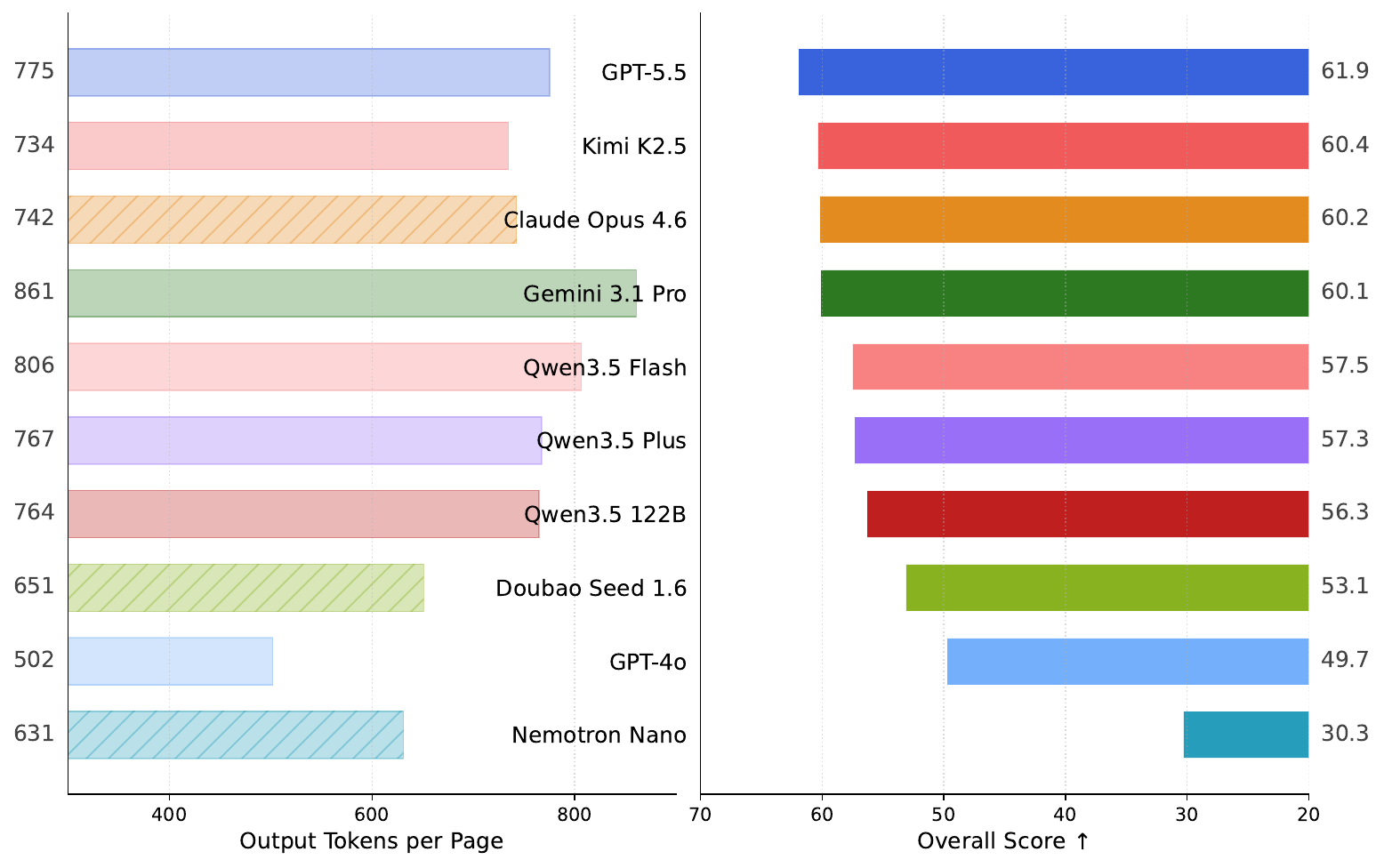}
    \caption{Token efficiency vs overall score.}
    \label{fig:token_efficiency}
\end{figure}

As each document spans $\sim$100 pages, we study how parsing and recognition performance change as the sliding window grows. We experiment with a sliding window from 1 to 15 pages. Figure~\ref{fig:scaling_window} shows that larger windows do not reliably improve long-document recognition. Reading order consistently degrades for nearly all models (e.g., edit distance: Claude 0.24 $\rightarrow$ 0.69; GPT-5.5: 0.16$\rightarrow$ 0.47 at $n=15$), since the model must order many more blocks while separating page-local contents from cross-page continuations. Text extraction and table parsing also worsen. Formula metrics are comparatively stable, suggesting that formulas are more spatially local (details in Appendix~\ref{app:sliding_window}).

\subsection{Token Efficiency}
\label{subsec:token_efficiency}
Figure~\ref{fig:token_efficiency} compares each model's average output-token cost against its overall parsing performance. Among the leading models, GPT-5.5 achieves the highest overall score with only a moderate output budget. Kimi K2.5 is more efficient: it recovers approximately 98\% of GPT-5.5's overall performance while using the lowest token budget among the top-performing group. In contrast, Gemini 3.1 Pro appears more token-hungry, paying a roughly 17\% token-cost more over Kimi K2.5 while yielding even lower accuracy. A closer inspection suggests that this additional generation often comes from redundant text, over-expanded structures, or less compact formatting rather than more faithful parsing. Nemotron Nano illustrates the opposite signal: despite its high output-token cost, it obtains the lowest overall score, indicating that longer outputs may reflect noisy extraction or poorly organized structure instead of better document understanding. Overall, these results suggests that the most effective parsers are those that produce compact, well-structured representations.

\begin{figure*}[htbp]
    \centering
    \begin{subfigure}[t]{0.18\linewidth}
        \includegraphics[width=\linewidth]{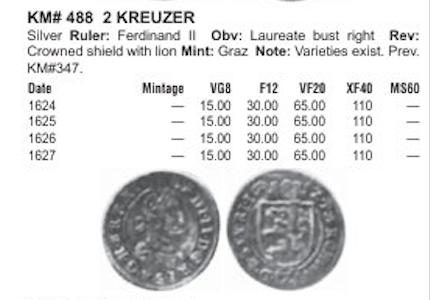}
        \subcaption{Source image}
        \label{fig:wireless_table_image}
    \end{subfigure}
    \hfill
    \begin{subfigure}[t]{0.18\linewidth}
        \includegraphics[width=\linewidth]{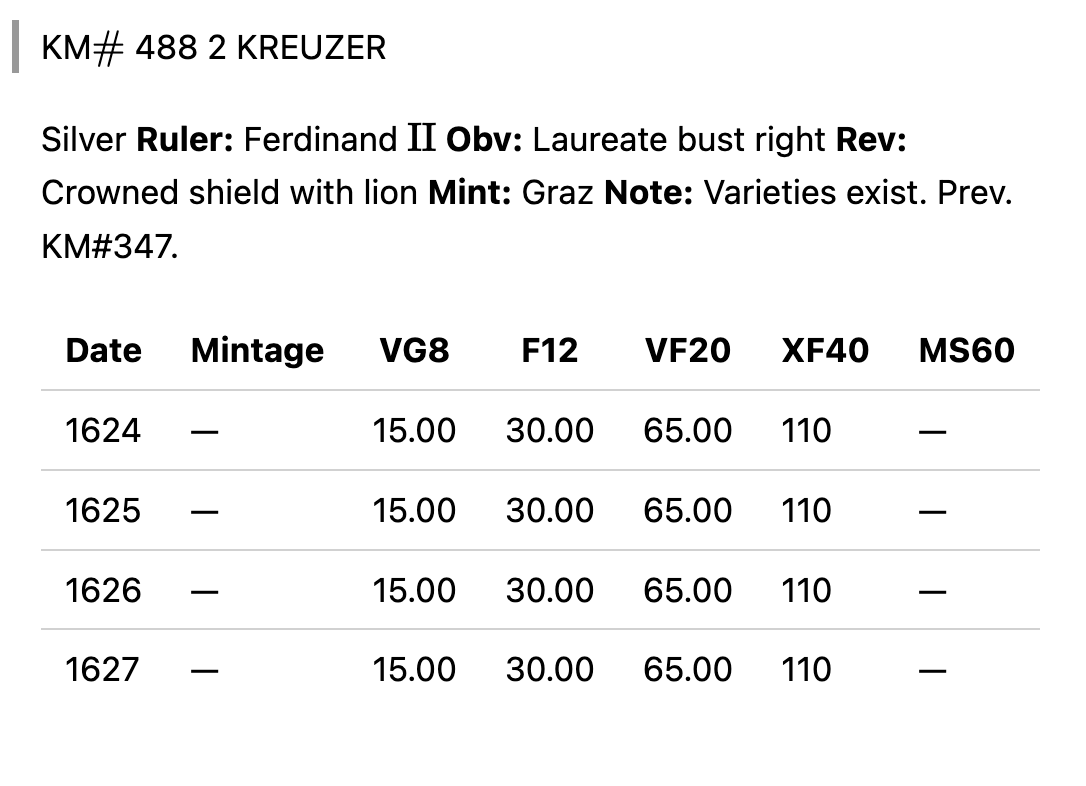}
        \subcaption{Ground-truth}
        \label{fig:rendered_gt_wireless}
    \end{subfigure}
    \hfill
    \begin{subfigure}[t]{0.18\linewidth}
        \includegraphics[width=\linewidth]{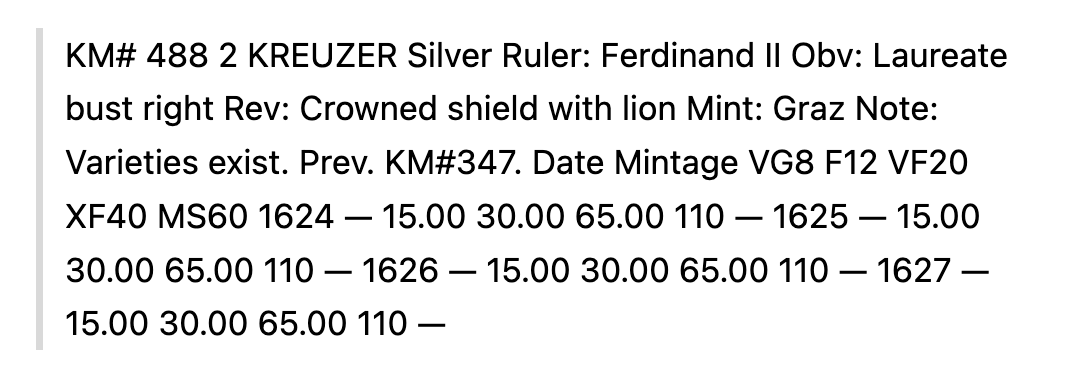}
        \subcaption{Kimi (curr.\ window)}
        \label{fig:rendered_kimi_wireless}
    \end{subfigure}
    \hfill
    \begin{subfigure}[t]{0.18\linewidth}
        \includegraphics[width=\linewidth]{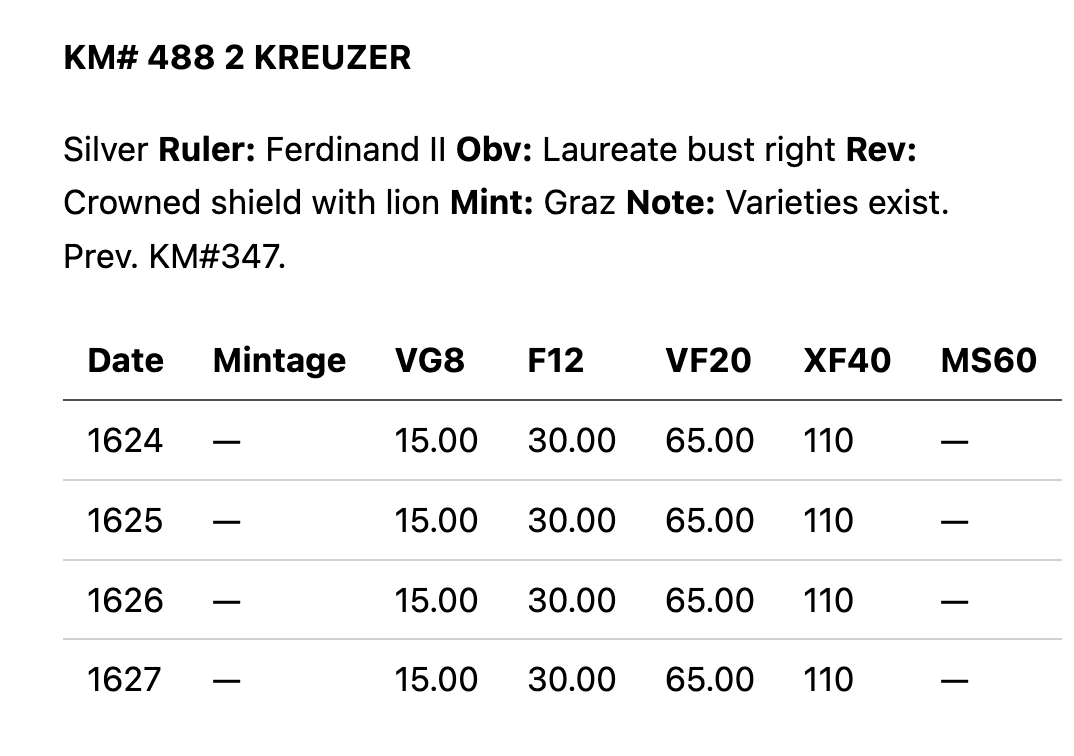}
        \subcaption{Kimi (prev.\ window)}
        \label{fig:rendered_kimi_prev_wireless}
    \end{subfigure}
    \hfill
    \begin{subfigure}[t]{0.18\linewidth}
        \includegraphics[width=\linewidth]{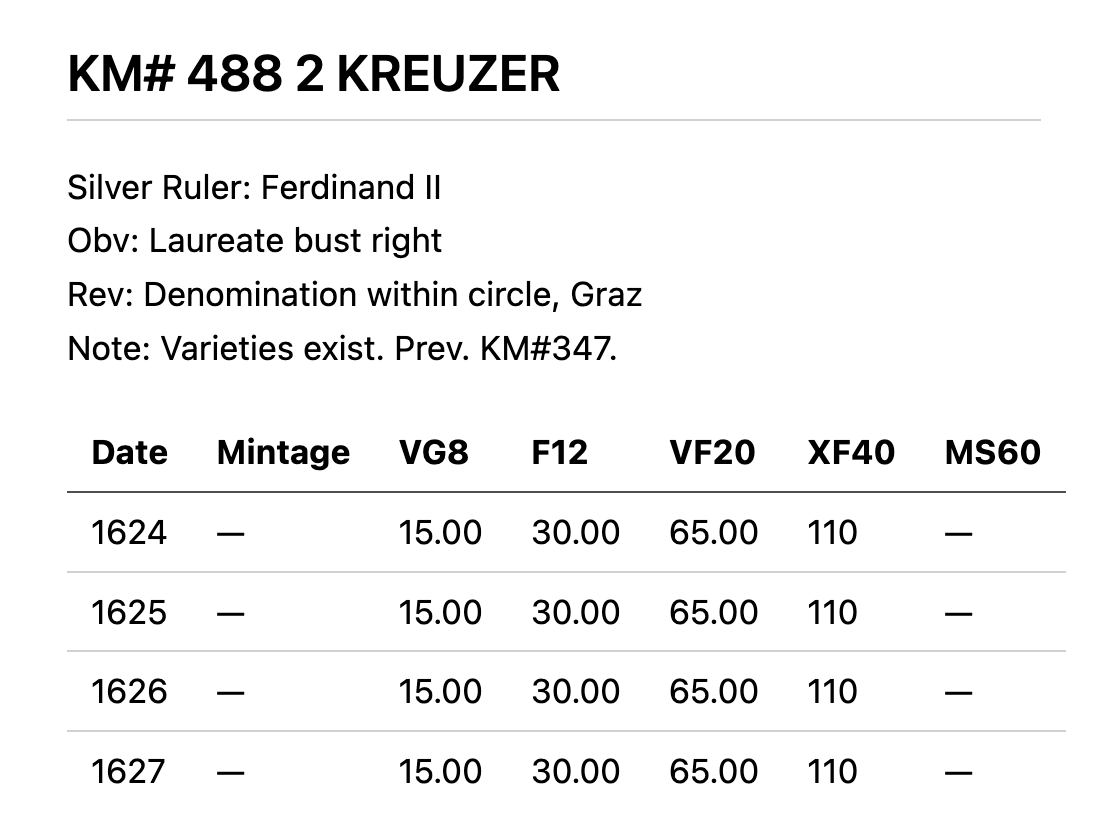}
        \subcaption{Doubao (curr.\ window)}
        \label{fig:rendered_doubao_wireless}
    \end{subfigure}
    \caption{Table case study. (\ref{fig:wireless_table_image}) source image, (\ref{fig:rendered_gt_wireless}) ground-truth HTML rendering. (\ref{fig:rendered_kimi_wireless}) Kimi in the current window collapses to plain text (\ref{fig:rendered_kimi_prev_wireless}). Kimi in the previous window produces correct content but ignores the HTML format instruction (\ref{fig:rendered_doubao_wireless}). Doubao attends almost entirely to prior-page content and misses the target table.}
    \label{fig:wireless_table_case_study}
\end{figure*}

\subsection{Case study: Table}
\label{subsec:case_study_table}
In Table~\ref{tab:combined-text-table}, we notice an interesting performance drop for wireless (borderless) in table recognition. From full line to wireless, Kimi drops from 49.1 $\rightarrow$ 21.4 TEDS; Doubao collapses to 8.1, which is near-random, revealing that most models rely heavily on visible grid lines for structure reconstruction. In this case study, we show an example of the borderless table as well as the rendered model output.

Figure~\ref{fig:wireless_table_case_study} shows the source table (\ref{fig:wireless_table_image}), the ground-truth HTML rendering (\ref{fig:rendered_gt_wireless}), and three model outputs. Doubao (\ref{fig:rendered_doubao_wireless}) collapses under page-positioning: in the previous window (where the table falls on the second page), it attends almost entirely to prior-page content, recovering 5 spurious tables while ignoring the 21 on the second page. In the current window (where the table is on the first page), Doubao correctly gets all 4 rows, confirming that the failure is positional, not an intrinsic weakness with borderless tables. Kimi (\ref{fig:rendered_kimi_wireless},~\ref{fig:rendered_kimi_prev_wireless}) ignores the HTML format instruction: in the previous window, it produces a correct markdown rendering, but in the current window, it collapses to plain text. 

\subsection{Case study: Optical Music Recognition}
\begin{figure}[ht!]
    \centering
    \includegraphics[width=\linewidth]{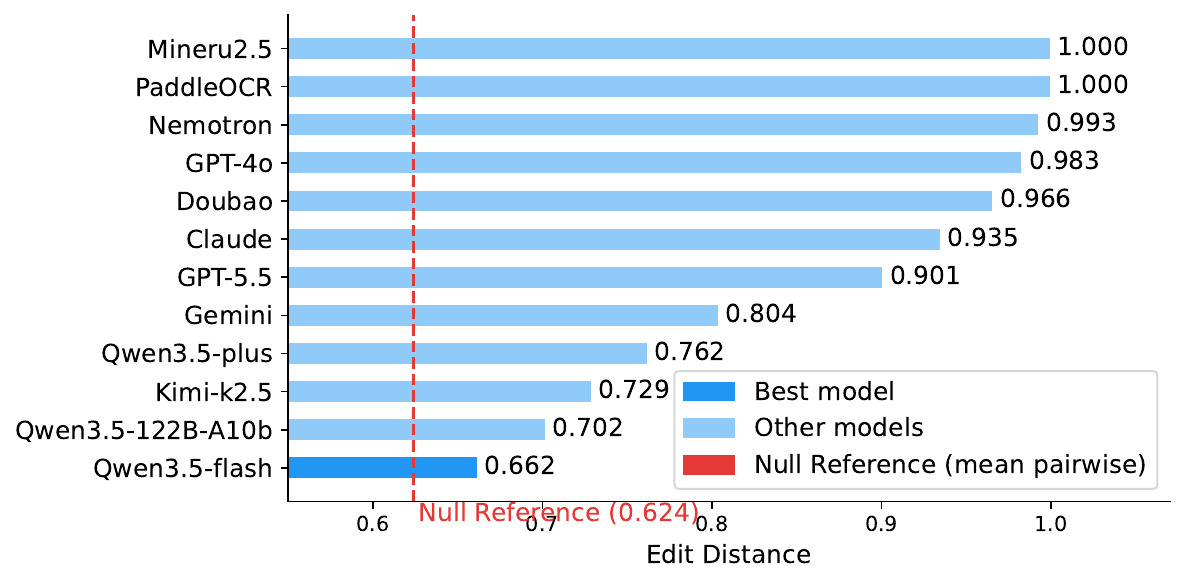}
    \caption{Music score $\rightarrow$ MusicXML transcription results (1-page window). Lower edit distance is better. The baseline is the mean pairwise edit distance across all 6 ground-truth documents.}
    \label{fig:music}
\end{figure}
The subject of Optical Music Recognition (OMR) is a brand-new subject not present in prior work~\citep{omnidocbench2024, ocrbench2024, fu2026ocrbench}. We are the first to evaluate this subject on a variety of frontier VLMs and specialized VLMs in the task of transcribing it to MusicXML. It exposes a failure mode that general document benchmarks do not: simultaneous visual comprehension of musical notation (clefs, noteheads, key/time signatures, articulations) and structured, schema-faithful XML generation. To anchor difficulty, we compute a cross-document null reference, the mean pairwise edit distance across all GT MusicXML files, yielding 0.624: the score expected from emitting an arbitrary unrelated score. As shown in Figure~\ref{fig:music}, no model beats this null, so as a schema-faithful transcription task, music remains unsolved for all systems evaluated. Pipeline OCR systems (MinerU-2.5, PaddleOCR) score 1.0 by design, since they cannot follow user instructions. For additional music results, see Appendix~\ref{app:music}.

\section{Conclusion}
\name{} is a difficulty-aware benchmark for expert-level document parsing, built from long-form multilingual books across 52 BISAC subject domains. Unlike prior OCR and document parsing benchmarks that mainly emphasize common genres or page-level recognition, \name{} targets challenging documents selected through parser-failure-based sampling and provides fine-grained annotations for layout, reading order, structural relations, and expert-domain content such as chemical diagrams, music notation, formulas, algorithms, and complex tables. Experiments on both specialized document parsers and frontier general-purpose VLMs show that strong performance on existing benchmarks does not reliably transfer to expert-level document parsing, with persistent failures on structurally complex subjects such as reference, design, medical, and music documents.
\clearpage
\section*{Ethical considerations}
\name{} is intended for research evaluation of document parsing systems. During dataset construction and quality control, we removed pages containing clearly sensitive personal information, private identifying information, or inappropriate content when identified, and annotators were instructed to flag such cases for review. The released annotations are intended for benchmarking and analysis, while the underlying source documents may remain subject to their original access, copyright, and licensing conditions. Users should follow the applicable terms of the original sources when using the benchmark.

\section*{Limitations}

While \name{} provides broad subject coverage and fine-grained annotations for difficult document parsing, it is designed primarily as a diagnostic benchmark for parsing and structural recognition rather than the full range of document-intelligence tasks. We use a unified inference prompt to ensure consistent comparison across models, but do not perform model-specific prompt engineering or extensive prompt optimization; the reported results therefore reflect standardized benchmark performance rather than the best achievable performance for each model. Certain specialized parsers, such as MinerU and PaddleOCR, do not favor user prompts or strictly follow format-specific instructions, so their scores on prompt-dependent outputs should be interpreted with caution and are reported mainly as reference baselines. The benchmark focuses on challenging pages selected from long-form book documents, which makes it especially useful for stress-testing current parsers on expert-level content; broader coverage of document sources, application scenarios, and downstream tasks remains an important direction.

Although BISAC subjects provide a useful organizing taxonomy, document difficulty also depends on fine-grained factors such as layout density, scan quality, visual artifacts, multilingual content, cross-page continuation, and domain-specific structures. We therefore view subject-level analysis as complementary to block-level and attribute-level analysis. In addition, expert-domain contents such as chemistry diagrams, music notation, formulas, algorithms, and complex tables are represented through structured formats such as normalized \LaTeX, HTML, and MusicXML. These formats enable scalable automatic evaluation, but they do not exhaustively cover all domain-specific knowledge representations or specialized package/library-dependent formats. Semantic, execution-based, or domain-specific metrics could further improve the evaluation of equivalent representations.

\bibliography{custom}

\clearpage
\appendix
\section*{Content of Appendix}
\begin{itemize}[noitemsep, topsep=0pt]
    \item[\ref{app:sourcing}] Data Sourcing and Pipeline
    \item[\ref{app:disagreement}] Multi-Parser Disagreement Score
    \item[\ref{app:schema}] Annotation Schema Reference
    \item[\ref{app:fullcat}] Full Per-Domain Statistics
    \item[\ref{app:language}] Language Coverage
    \item[\ref{app:abbrev}] Subject Name Abbreviations
    \item[\ref{app:inference_prompt}] Inference Prompt
    \item[\ref{app:additional_results}] Additional Results
\end{itemize}

\section{Data Sourcing and Pipeline}
\label{app:sourcing}

The seed corpus is sourced from 
a publicly accessible digital library aggregating trade and academic
books with peer-reviewed research journals across a wide range of
professional and consumer domains. Every document is assigned to the
BISAC 2024 subject taxonomy and \emph{human-verified} by a trained
reviewer; where multiple BISAC codes apply, the primary code is kept.

After classification we apply difficulty-aware sampling
(Appendix~\ref{app:disagreement}) and take the top-$k$ books per BISAC
subject, allocating exactly 6 PDFs per subject so that no single
subject can dominate aggregate scores.

\paragraph{License.} The \name{} annotations are released under the
Creative Commons CC0 1.0 Universal license (public domain dedication).

\paragraph{Annotators and Compensation.} The released \name{} labels were produced by a team of annotators recruited under the mixed workflow described in Sec.~\ref{sec:quality_control}. General document elements (text blocks, titles, simple tables, reading order, and structural relations) were handled by trained annotators holding at least an undergraduate degree, while expert-content blocks were routed to domain specialists with MS- or PhD-level training in the relevant field: chemistry-majored annotators for molecular structures and reaction schemes, music-literate annotators for MusicXML transcription of scores, and LaTeX-fluent mathematicians for dense-formula pages. Annotators were compensated per annotated block rather than at a fixed hourly or per-page rate, with the per-unit price tiered by block complexity so that domain-expert blocks (chemistry, music, dense formulas, complex tables) paid materially more per unit than plain text or simple structural elements. All compensation was at or above the prevailing local market rate for skilled annotation work.

\begin{figure*}[h]
  \centering
  \includegraphics[width=\linewidth]{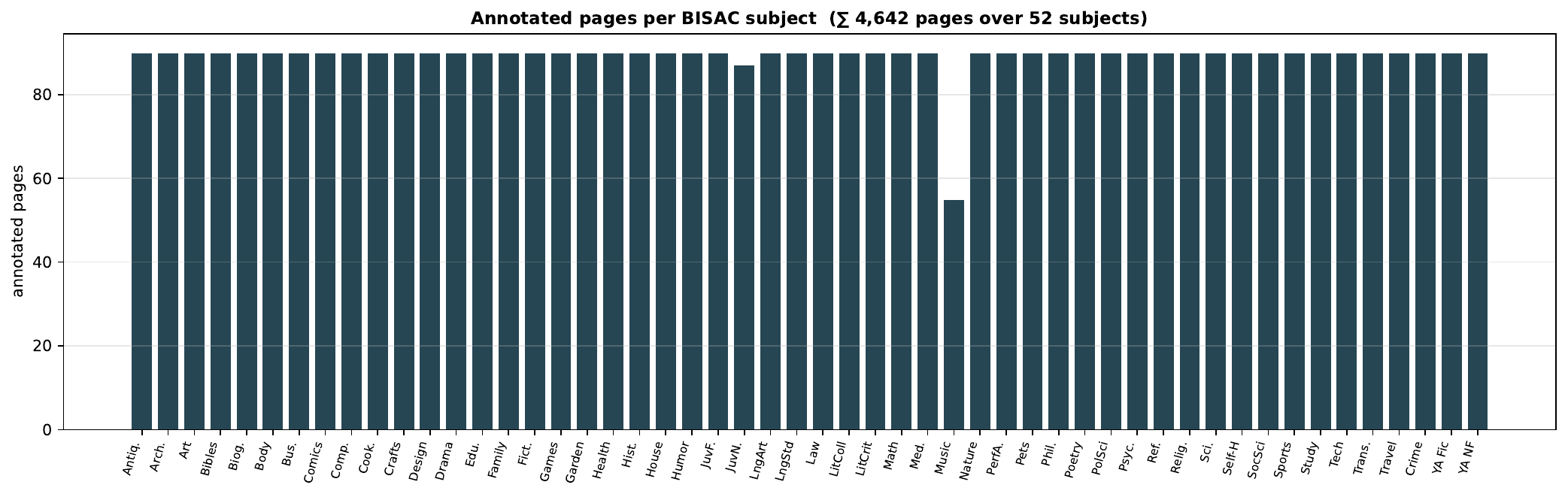}
  \caption{Number of annotated pages per BISAC subject in the full
  \name{} release. Each subject contains 6 source PDFs, for a total of
  4{,}514 annotated pages. The distribution reflects the natural variation
  in document length after applying our difficulty-aware sampling process.}
  \label{fig:anno-pages}
\end{figure*}
\begin{figure*}[h]
  \centering
  \includegraphics[width=\linewidth]{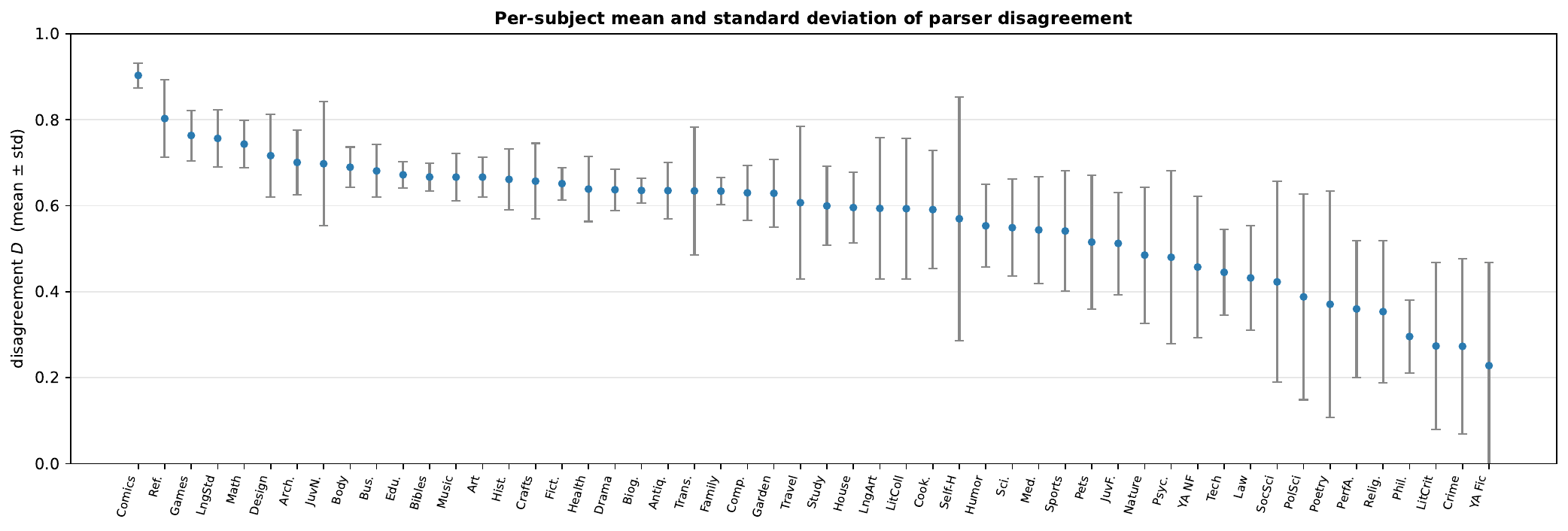}
  \caption{Mean and standard deviation of four-parser disagreement for
  each BISAC subject. The mean captures the overall parsing difficulty of
  a subject, while the standard deviation reflects within-subject
  heterogeneity across sampled documents. Subjects with both high mean and
  high variance, such as \textsc{Design} and \textsc{Reference}, serve as
  particularly discriminative stress tests because they contain diverse
  layouts and failure modes.}
  \label{fig:avg-std}
\end{figure*}

\section{Multi-Parser Disagreement Score}
\label{app:disagreement}

To bias annotation toward pages that current parsers cannot already
solve, we score each candidate book by the disagreement of three
frontier parsers---PaddleOCR-VL-1.5, MinerU-VLM, and Gemini-3.1-pro-preview---on the
Markdown transcription of all pages. Let $T_{m,b}$ denote the
concatenation of model-$m$ Markdown outputs over book $b$. The
text-disagreement term is the mean normalized Levenshtein distance
over the six parser pairs,
\begin{equation}
D^{\text{text}}_b = \frac{1}{6}\!\!\sum_{(i,j)}\!\frac{\mathrm{Lev}(T_{m_i,b}, T_{m_j,b})}{\max(|T_{m_i,b}|,|T_{m_j,b}|)} .
\label{eq:disagreement-text}
\end{equation}
A composite score
$D_b = \alpha D^{\text{text}}_b + \gamma D^{\text{struct}}_b + \delta D^{\text{conf}}_b$
optionally adds a structural-tag-mismatch term $D^{\text{struct}}_b$
(headings, list items, table cells) and a low-confidence penalty
$D^{\text{conf}}_b$ for empty/malformed outputs. In our release we use
$\alpha = 0.6$, $\gamma = 0.3$, $\delta = 0.1$ and admit books with
$D_b > \tau = 0.5$ into the annotation queue.

\section{Annotation Schema Reference}
\label{app:schema}

\subsection{OmniDocJSON Page Record}
{\small
\begin{verbatim}
{
  "page_info": {
    "page_no": 7,
    "height": 1500, "width": 2000,
    "image_path": "<file>.jpg",
    "page_attribute": {
      "subject": "<BISAC code>",
      "challenge_type": "perception" |
        "structural_reconstruction" |
        "domain_reasoning",
      "data_source": "academic_literature",
      "language": "english",
      "layout": "double_column",
      "special_issue": ["table_full_line"]
    }
  },
  "layout_dets": [ ... ],
  "extra": { "relation": [ ... ] }
}
\end{verbatim}
}

\subsection{Page-Level Attributes}
\begin{description}\setlength\itemsep{2pt}
  \item[\textit{subject}.] One of 52 BISAC top-level domains.
  \item[\textit{challenge\_type}.] Perception, structural reconstruction,
        or domain reasoning.
  \item[\textit{data\_source}.] One of nine genres: academic literature, book,
        colorful textbook, exam paper, note, magazine, research report,
        newspaper, and PPT-to-PDF.
  \item[\textit{language}.] English, simplified Chinese, en--ch mixed, or other.
  \item[\textit{layout}.] Single-, double-, three-, or 1-and-more-column,
        plus an ``other'' bucket.
  \item[\textit{special\_issue} (multi-select).] e.g.\ colorful background,
        horizontal table, table span, table with formula, fuzzy scan.
\end{description}

\subsection{Block-Level Categories}
Quadrilateral block annotations fall into five groups:
\begin{description}\setlength\itemsep{2pt}
  \item[Textual blocks.] Title, text block, reference, and the caption /
        footnote / explanation variants attached to figures, tables,
        equations, and code (including \textit{code\_txt} and
        \textit{code\_algorithm} families).
  \item[Page-element blocks (no reading order).] Header, footer, page number,
        and page footnote.
  \item[Tables.] LaTeX-annotated, described by \textit{table\_layout},
        \textit{line}, \textit{language}, \textit{with\_span},
        \textit{include\_background}, \textit{include\_equation},
        \textit{include\_photo}, and \textit{with\_structured\_text}.
  \item[Formulas.] Isolated equations with \textit{formula\_type} $\in$
        \{print, handwritten\}; chemical structures are flagged via
        \textit{need\_mask}.
  \item[Reading-order-only blocks.] Figure, table mask, text mask.
  \item[No-property blocks.] Ignore-formula/abandon.
\end{description}

\subsection{Inter-Block Relations}
Each record in \textit{extra.relation} has the form
$(\textit{source\_anno\_id},\ \textit{target\_anno\_id},\ \textit{relation\_type})$.
Two relation types are defined:
\begin{description}\setlength\itemsep{2pt}
  \item[\emph{parent--son}.] Links a figure, table, equation, or code
        parent to its caption or footnote child.
  \item[\emph{truncated}.] Marks two blocks that form one logical unit;
        for cross-page truncation, \textit{target\_anno\_id} is set to $-1$.
\end{description}
A \textit{merge\_list} attribute on a parent text block stores the original
child blocks (e.g.\ TOC entries), while the parent's \textit{text} field
carries the flattened content.

\section{Full Per-Domain Statistics}
\label{app:fullcat}

To provide a complete view of the benchmark composition and sampling
difficulty, we report per-domain statistics for all 52 BISAC subjects.
These statistics complement the aggregate analysis in the main paper by
showing how annotated pages and parser-disagreement scores vary across
subjects.

Together, these statistics show that \name{} is not only broad in subject
coverage, but also heterogeneous in parsing difficulty. Some domains are
consistently challenging across parsers, while others exhibit large
within-domain variation, suggesting that expert-document parsing requires
robustness to both domain-specific visual conventions and document-level
layout diversity. The complete 52-row per-subject table, including PDF
counts, annotated page counts, and mean, minimum, maximum, and standard
deviation of parser disagreement.

\section{Language Coverage}
\label{app:language}

\begin{table}[!htbp]
\centering
\small
\setlength{\tabcolsep}{6pt}
\begin{tabular}{lrr}
\toprule
\textbf{Language} & \textbf{Pages} & \textbf{Share} \\
\midrule
English                  & 3{,}083 & 68.30\% \\
Other                    &    518  & 11.48\% \\
Russian                  &    334  &  7.40\% \\
Simplified Chinese       &    152  &  3.37\% \\
German                   &    102  &  2.26\% \\
Japanese                 &     74  &  1.64\% \\
French                   &     72  &  1.60\% \\
Turkish                  &     46  &  1.02\% \\
Spanish                  &     45  &  1.00\% \\
Traditional Chinese      &     41  &  0.91\% \\
English--Chinese mixed   &     20  &  0.44\% \\
Portuguese               &     15  &  0.33\% \\
Egyptian Arabic          &      9  &  0.20\% \\
Arabic                   &      3  &  0.07\% \\
\midrule
\textbf{Total}           & \textbf{4{,}514} & \textbf{100\%} \\
\bottomrule
\end{tabular}
\caption{Per-page primary-language distribution across
\name. \emph{Other} aggregates pages
whose primary language was tagged as ``other'' by the annotator.}
\label{tab:language}
\end{table}

Every page in \name carries a per-page \texttt{language}
attribute. Tab.~\ref{tab:language} reports the page counts across all
fourteen distinct languages present in the release. English accounts
for roughly two-thirds of pages, but the remaining $\sim$32\% spans
a long tail of major world languages (Russian, Simplified and
Traditional Chinese, German, Japanese, French, Turkish, Spanish,
Portuguese, and Arabic) which makes the benchmark multilingual.

\section{Subject Name Abbreviations}
\label{app:abbrev}

To save horizontal space, BISAC subject domains are abbreviated in the
main-body tables and figures. The mapping below lists every
abbreviation used together with its full BISAC name:
\textbf{Antiq} (Antiques \& Collectibles); \textbf{Arch}
(Architecture); \textbf{Biog} (Biography \& Autobiography);
\textbf{Comics} (Comics \& Graphic Novels); \textbf{Crafts} (Crafts \&
Hobbies); \textbf{Des} (Design); \textbf{Edu} (Education);
\textbf{Fam} (Family \& Relationships); \textbf{Fic} (Fiction);
\textbf{Games} (Games \& Activities); \textbf{Hist} (History);
\textbf{Lit Crit} (Literary Criticism); \textbf{Med} (Medical);
\textbf{Poet} (Poetry); \textbf{Psych} (Psychology); \textbf{Ref}
(Reference); \textbf{Soc Sci} (Social Science); \textbf{Study}
(Study Aids); \textbf{YA Fic} (Young Adult Fiction).

\section{Inference Prompt}
\label{app:inference_prompt}
\onecolumn

\begin{figure}[p]
\centering

\begin{tcolorbox}[
    colback=myblue,
    colframe=myframe,
    title=Unified Inference Prompt (1/2),
    breakable
]

\begin{Verbatim}[
    fontsize=\small,
    breaklines=true,
    baselinestretch=0.88
]
You are an AI assistant specialized in converting document images to Markdown format. You are given {num_images} consecutive page image(s). Convert all pages in order into a single continuous Markdown document. Please follow these instructions for the conversion:

1. Text Processing:
   - Accurately recognize all text content in the image without guessing or inferring.
   - Transcribe text in its original language and script — do not translate. This includes but is not limited to Chinese, English, Japanese, Russian, Arabic, and other languages.
   - For degraded, blurry, or noisy scans: transcribe what is legible; do not invent or guess characters that are not clearly readable.
   - Maintain the original document structure, including headings, paragraphs, lists, etc.
   - For multi-column layouts, output columns in natural reading order (left-to-right, top-to-bottom).

2. Mathematical Formula Processing:
   - Convert all mathematical formulas to LaTeX format.
   - Enclose inline formulas with \( \). For example: This is an inline formula \( E = mc^2 \)
   - Enclose block formulas with \[ \]. For example: \[ \frac{-b \pm \sqrt{b^2 - 4ac}}{2a} \]

3. Chemical Notation:
   - For chemical reaction equations and ionic equations, use the \ce{} syntax from the mhchem package.
     Example inline: \( \ce{H2O} \)
     Example block: \[ \ce{2H2 + O2 -> 2H2O} \]
   - For 2D molecular structures given as SMILES strings, preserve the SMILES notation exactly inside a smiles block:
     ```smiles
     CCO
     ```
   - For chemical symbols, element names, subscripts, and superscripts that appear in running text (e.g. H$_2$O, CO$_2$), render them using \ce{} or standard LaTeX subscript/superscript notation.
   - Preserve bond notation (single, double, triple), stereochemical indicators, reaction arrows, catalyst labels, and condition markers exactly as they appear.

4. Domain-Specific Notation:
   - Legal documents: preserve clause and sub-clause numbering exactly (e.g. Article 3.2(a)(i)).
   - Biomedical text: preserve gene names, protein names, dosage units, and clinical notation without alteration.
   - Financial documents: preserve ticker symbols, currency symbols, and numerical formats (e.g. $1,234.56, €, ¥).
   - Code and pseudocode: wrap in fenced code blocks with the appropriate language tag.
     Example:
     ```python
     def foo():
         pass
     ```
   - Diagram labels, callout text, and annotation text that is legible within a figure: transcribe as a plain text paragraph immediately after the figure's position; do not describe the diagram itself.

5. Table Processing:
   - Convert tables to HTML format.
   - Wrap the entire table with <table> and </table>.
   - For long tables that span multiple pages, output all rows in a single continuous HTML table.
   - For tables with merged cells, use rowspan and colspan attributes.

6. Comics and Dialog Balloons:
   - Extract all readable text in visual reading order: speech bubbles, thought bubbles, caption boxes, narration boxes, and any text visible within panels.
   - Output each piece of extracted text as a separate paragraph.
   - Do not add speaker attributions, panel labels, or separators. Do not output placeholder text for panels with no text.

7. Forms:
   - Render form fields and their labels as key–value pairs using a definition list or plain text, preserving the original field order.
   - For filled-in values, transcribe the handwritten or typed content exactly.
   - For checkboxes or radio buttons, indicate the selection state: [x] for checked, [ ] for unchecked.
\end{Verbatim}

\end{tcolorbox}

\caption{Unified inference prompt used in our evaluation (Part 1).}
\label{fig:unified_prompt_part1}
\end{figure}

\begin{figure}[p]
\centering

\begin{tcolorbox}[
    colback=myblue,
    colframe=myframe,
    title=Unified Inference Prompt (2/2),
    breakable
]

\begin{Verbatim}[
    fontsize=\small,
    breaklines=true,
    baselinestretch=0.88
]

8. Figure Handling:
   - Ignore purely visual figures (photographs, illustrations, charts) where no text is embedded. Do not describe or caption them.
   - Exception: extract any embedded text labels, axis labels, legend text, or annotations that are legible (see rule 4 above).
   
9. Music:
   - If the image contains a music score (sheet music with staves, notes, clefs, time signatures, etc.), render it in MusicXML format inside a fenced code block:
     ```musicxml
     <part-list>
       <score-part id="P1">
         <part-name>...</part-name>
         <part-abbreviation>...</part-abbreviation>
       </score-part>
     </part-list>
     <part id="P1">
       <measure number="1">
         ...
       </measure>
     </part>
     ```
   - Do not include an XML declaration (`<?xml ... ?>`) or `<score-partwise>` wrapper.
   - Include `<part-list>` with `<part-name>` and `<part-abbreviation>` for any instrument names or abbreviations visible in the score margin.
   - Transcribe all visible elements: clef, key signature, time signature, notes, rests, ties, slurs, dynamics, tempo markings, articulations, lyrics, stave count, rehearsal marks, repeat signs, barline types, ottava markings (8va/8vb), fermatas, trills, and other ornament symbols.
   - Key signatures must use the `<fifths>` integer encoding: positive = sharps (e.g. `<fifths>1</fifths>` = G major), negative = flats, zero = C major.
   - For keyboard instruments or any score with multiple staves per part, use `<staves>` and add a `<staff>` number attribute to each note to indicate which staff it belongs to.
   - Omit all non-visual metadata that cannot be read from the image: `<print>`, `<system-layout>`, `<system-margins>`, `<staff-layout>`, pixel/point measurements, `<divisions>`, `<midi-device>`, `<midi-instrument>`, and any rendering coordinates.
   - For text-only music content (song lyrics without notation, chord charts, tablature in plain text), transcribe as regular Markdown text, preserving chord symbols above lyrics lines.

10. Output Format:
   - Ensure the output Markdown document has a clear structure with appropriate line breaks between elements.
   - For complex layouts, maintain the original document's reading order as closely as possible.
   - Do not wrap the entire output in a code block. Do not add explanations, comments, or meta-commentary.
\end{Verbatim}

\end{tcolorbox}

\caption{Unified inference prompt used in our evaluation (Part 2).}
\label{fig:unified_prompt_part2}
\end{figure}

\twocolumn
The full inference prompt is in Figure~\ref{fig:unified_prompt_part1} and Figure~\ref{fig:unified_prompt_part2}.

\section{Additional Results}
\label{app:additional_results}

The top four general VLMs (GPT-5.5, Kimi, Claude, Gemini) cluster within 1.8 overall points, with no single model dominant across all metrics; GPT-5.5 leads on text and reading order, while Kimi leads on tables.

\paragraph{GPT-5.5 vs.\ GPT-4o.}
GPT-5.5 (61.94) outperforms GPT-4o (49.73) by 12.2 points overall, with the largest gains on text\_Edit (0.189 vs.\ 0.366) and formula\_Edit (0.359 vs.\ 0.468).

\paragraph{Formula and table metrics.}
For Kimi, CDM and edit distance diverge: it records the lowest formula\_CDM (27.04) among frontier models while remaining competitive on formula\_Edit (0.350), indicating that the two metrics capture different aspects of formula quality. On tables, MinerU 2.5 (TEDS 55.85, TEDS\_S 63.70) outperforms all general VLMs despite not supporting user prompts.

\paragraph{Prompt-incapable systems and outliers.}
Since MinerU 2.5 and PaddleOCR do not support user prompts, their reading order and text scores are not interpretable as capability measures. PaddleOCR's formula\_Edit (0.415) is further inflated by a delimiter mismatch (\verb|$$...$$| vs.\ ground truth \verb|\[...\]|). Nemotron-Nano-12B (30.33) falls 19 points below the next weakest general VLM (GPT-4o), making it a clear outlier.

\subsection{Per-subject breakdown}
\begin{table*}[htbp!]
\centering
\setlength{\tabcolsep}{3pt}
\resizebox{\linewidth}{!}{
\begin{tabular}{lrrrrrrrrrrrr|r}
\toprule
Subject & Claude & Doubao & Gemini & GPT-4o & GPT-5.5 & Kimi & MinerU & Nemotron & PaddleOCR & Qwen-122B & Qwen-Flash & Qwen-Plus & Avg \\
\midrule
Antiques \& Collectibles            & 0.299 & 0.593 & 0.302 & 0.666 & \textbf{0.173} & 0.414 & 0.658 & 0.870 & 0.885 & 0.354 & 0.390 & 0.345 & \cellcolor[rgb]{0.87,0.86,0.78} 0.496 \\
Architecture                        & 0.208 & 0.264 & 0.133 & 0.403 & 0.153 & \textbf{0.095} & 0.206 & 0.591 & 0.715 & 0.123 & 0.212 & 0.140 & \cellcolor[rgb]{0.80,0.91,0.78} 0.270 \\
Art                                 & 0.311 & 0.279 & 0.192 & 0.336 & 0.196 & 0.193 & 0.355 & 0.485 & 0.790 & 0.201 & 0.219 & \textbf{0.188} & \cellcolor[rgb]{0.82,0.90,0.78} 0.312 \\
Bibles                              & 0.203 & 0.374 & 0.192 & 0.408 & 0.205 & \textbf{0.182} & 0.329 & 0.763 & 0.612 & 0.215 & 0.214 & 0.237 & \cellcolor[rgb]{0.82,0.90,0.78} 0.328 \\
Biography \& Autobiography          & \textbf{0.001} & 0.111 & 0.103 & 0.110 & 0.092 & 0.102 & 0.118 & 0.506 & 0.735 & 0.102 & 0.133 & 0.118 & \cellcolor[rgb]{0.78,0.93,0.78} 0.186 \\
Body, Mind \& Spirit                & 0.510 & 0.203 & 0.159 & 0.445 & 0.218 & \textbf{0.094} & 0.150 & 0.713 & 0.494 & 0.122 & 0.170 & 0.181 & \cellcolor[rgb]{0.81,0.91,0.78} 0.288 \\
Business \& Economics               & 0.077 & 0.271 & 0.133 & 0.337 & 0.089 & \textbf{0.028} & 0.098 & 0.661 & 0.853 & 0.181 & 0.132 & 0.165 & \cellcolor[rgb]{0.80,0.92,0.78} 0.252 \\
Comics \& Graphic Novels            & 0.176 & 0.320 & \textbf{0.152} & 0.866 & 0.198 & 0.278 & 0.962 & 0.761 & 0.724 & 0.277 & 0.258 & 0.157 & \cellcolor[rgb]{0.85,0.88,0.78} 0.427 \\
Computers                           & 0.357 & 0.401 & 0.239 & 0.332 & 0.201 & \textbf{0.189} & 0.399 & 0.692 & 0.828 & 0.294 & 0.284 & 0.277 & \cellcolor[rgb]{0.83,0.89,0.78} 0.374 \\
Cooking                             & 0.198 & 0.391 & 0.434 & 0.300 & \textbf{0.075} & 0.107 & 0.171 & 0.655 & 0.834 & 0.345 & 0.413 & 0.515 & \cellcolor[rgb]{0.83,0.89,0.78} 0.370 \\
Crafts \& Hobbies                   & \textbf{0.174} & 0.663 & 0.358 & 0.583 & 0.245 & 0.212 & 0.511 & 0.814 & 0.863 & 0.397 & 0.427 & 0.434 & \cellcolor[rgb]{0.86,0.87,0.78} 0.473 \\
Design                              & 0.634 & 0.647 & \textbf{0.423} & 0.694 & 0.532 & 0.620 & 0.593 & 0.788 & 0.710 & 0.619 & 0.603 & 0.549 & \cellcolor[rgb]{0.90,0.84,0.78} 0.618 \\
Drama                               & \textbf{0.076} & 0.414 & 0.339 & 0.345 & 0.098 & 0.110 & 0.092 & 0.548 & 0.871 & 0.263 & 0.336 & 0.451 & \cellcolor[rgb]{0.82,0.90,0.78} 0.329 \\
Education                           & 0.226 & 0.316 & 0.141 & 0.272 & 0.114 & \textbf{0.083} & 0.136 & 0.463 & 0.868 & 0.153 & 0.180 & 0.175 & \cellcolor[rgb]{0.80,0.91,0.78} 0.261 \\
Family \& Relationships             & 0.161 & 0.270 & 0.163 & 0.306 & 0.136 & 0.143 & \textbf{0.130} & 0.385 & 0.764 & 0.214 & 0.225 & 0.160 & \cellcolor[rgb]{0.80,0.91,0.78} 0.255 \\
Fiction                             & 0.112 & 0.185 & 0.130 & 0.151 & 0.092 & \textbf{0.039} & 0.092 & 0.601 & 0.703 & 0.088 & 0.228 & 0.208 & \cellcolor[rgb]{0.79,0.92,0.78} 0.219 \\
Games \& Activities                 & 0.341 & 0.639 & \textbf{0.340} & 0.654 & 0.384 & 0.375 & 0.580 & 0.795 & 0.804 & 0.465 & 0.504 & 0.435 & \cellcolor[rgb]{0.87,0.86,0.78} 0.526 \\
Gardening                           & 0.087 & 0.597 & 0.107 & 0.530 & 0.110 & \textbf{0.078} & 0.140 & 0.671 & 0.798 & 0.122 & 0.270 & 0.258 & \cellcolor[rgb]{0.82,0.90,0.78} 0.314 \\
Health \& Fitness                   & 0.053 & 0.421 & 0.152 & 0.385 & 0.070 & \textbf{0.035} & 0.128 & 0.739 & 0.782 & 0.119 & 0.162 & 0.271 & \cellcolor[rgb]{0.81,0.91,0.78} 0.276 \\
History                             & \textbf{0.177} & 0.538 & 0.293 & 0.497 & 0.207 & 0.187 & 0.364 & 0.743 & 0.644 & 0.186 & 0.300 & 0.242 & \cellcolor[rgb]{0.83,0.89,0.78} 0.365 \\
House \& Home                       & \textbf{0.097} & 0.531 & 0.122 & 0.432 & 0.100 & 0.108 & 0.181 & 0.710 & 0.716 & 0.151 & 0.179 & 0.229 & \cellcolor[rgb]{0.81,0.91,0.78} 0.296 \\
Humor                               & 0.105 & 0.174 & 0.152 & 0.239 & 0.138 & 0.247 & 0.207 & 0.621 & 0.626 & \textbf{0.096} & 0.191 & 0.184 & \cellcolor[rgb]{0.80,0.92,0.78} 0.248 \\
Juvenile Fiction                    & \textbf{0.105} & 0.257 & 0.152 & 0.342 & 0.135 & 0.151 & 0.544 & 0.548 & 0.572 & 0.146 & 0.162 & 0.126 & \cellcolor[rgb]{0.80,0.91,0.78} 0.270 \\
Juvenile Nonfiction                 & 0.230 & 0.396 & \textbf{0.189} & 0.398 & 0.310 & 0.295 & 0.469 & 0.605 & 0.777 & 0.338 & 0.442 & 0.368 & \cellcolor[rgb]{0.84,0.88,0.78} 0.401 \\
Language Arts \& Disciplines        & 0.645 & 0.693 & 0.704 & 0.645 & \textbf{0.633} & 0.682 & 0.774 & 0.816 & 0.916 & 0.651 & 0.673 & 0.663 & \cellcolor[rgb]{0.93,0.82,0.78} 0.708 \\
Language Study                      & 0.402 & 0.495 & \textbf{0.315} & 0.624 & 0.363 & 0.343 & 0.630 & 0.920 & 0.644 & 0.365 & 0.440 & 0.355 & \cellcolor[rgb]{0.86,0.86,0.78} 0.491 \\
Law                                 & \textbf{0.277} & 0.470 & 0.374 & 0.500 & 0.383 & 0.420 & 0.649 & 0.789 & 0.630 & 0.345 & 0.483 & 0.366 & \cellcolor[rgb]{0.86,0.87,0.78} 0.474 \\
Literary Collections                & 0.256 & 0.322 & 0.182 & 0.496 & 0.212 & \textbf{0.156} & 0.386 & 0.808 & 0.719 & 0.236 & 0.258 & 0.289 & \cellcolor[rgb]{0.83,0.89,0.78} 0.360 \\
Literary Criticism                  & \textbf{0.131} & 0.258 & 0.144 & 0.229 & 0.190 & 0.140 & 0.271 & 0.662 & 0.651 & 0.178 & 0.186 & 0.274 & \cellcolor[rgb]{0.81,0.91,0.78} 0.276 \\
Mathematics                         & 0.234 & 0.439 & 0.251 & 0.333 & \textbf{0.216} & 0.288 & 0.270 & 0.688 & 0.700 & 0.263 & 0.264 & 0.236 & \cellcolor[rgb]{0.82,0.89,0.78} 0.348 \\
Medical                             & 0.261 & 0.544 & 0.271 & 0.542 & \textbf{0.187} & 0.515 & 0.552 & 0.772 & 0.827 & 0.542 & 0.524 & 0.506 & \cellcolor[rgb]{0.87,0.86,0.78} 0.503 \\
Music$^\dagger$                     & 0.935 & 0.966 & 0.804 & 0.983 & 0.901 & 0.729 & 1.000 & 0.993 & 1.000 & 0.702 & \textbf{0.662} & 0.762 & \cellcolor[rgb]{0.97,0.78,0.78} 0.870 \\
Nature                              & 0.306 & 0.473 & 0.326 & 0.374 & 0.278 & \textbf{0.244} & 0.576 & 0.685 & 0.708 & 0.373 & 0.352 & 0.296 & \cellcolor[rgb]{0.84,0.88,0.78} 0.416 \\
Performing Arts                     & \textbf{0.049} & 0.325 & 0.166 & 0.353 & 0.081 & 0.143 & 0.141 & 0.687 & 0.798 & 0.169 & 0.206 & 0.206 & \cellcolor[rgb]{0.81,0.91,0.78} 0.277 \\
Pets                                & \textbf{0.083} & 0.339 & 0.152 & 0.214 & 0.113 & 0.129 & 0.199 & 0.754 & 0.840 & 0.153 & 0.159 & 0.211 & \cellcolor[rgb]{0.81,0.91,0.78} 0.279 \\
Philosophy                          & 0.212 & 0.269 & 0.212 & 0.263 & 0.160 & \textbf{0.133} & 0.203 & 0.633 & 0.753 & 0.197 & 0.160 & 0.209 & \cellcolor[rgb]{0.81,0.91,0.78} 0.284 \\
Poetry                              & 0.076 & 0.171 & 0.234 & 0.174 & 0.129 & \textbf{0.069} & 0.221 & 0.582 & 0.618 & 0.125 & 0.218 & 0.256 & \cellcolor[rgb]{0.79,0.92,0.78} 0.239 \\
Political Science                   & 0.209 & 0.279 & 0.202 & 0.474 & 0.198 & 0.157 & 0.249 & 0.741 & 0.610 & 0.214 & \textbf{0.144} & 0.215 & \cellcolor[rgb]{0.81,0.90,0.78} 0.308 \\
Psychology                          & 0.111 & 0.400 & 0.173 & 0.332 & 0.185 & \textbf{0.110} & 0.211 & 0.753 & 0.809 & 0.138 & 0.158 & 0.171 & \cellcolor[rgb]{0.81,0.91,0.78} 0.296 \\
Reference                           & 0.833 & 0.909 & 0.860 & 0.906 & 0.788 & 0.819 & 0.769 & 0.926 & 0.938 & \textbf{0.768} & 0.840 & 0.883 & \cellcolor[rgb]{0.97,0.78,0.78} 0.853 \\
Religion                            & 0.174 & 0.340 & 0.168 & 0.427 & 0.216 & 0.144 & 0.229 & 0.719 & 0.684 & 0.194 & 0.150 & \textbf{0.133} & \cellcolor[rgb]{0.81,0.91,0.78} 0.298 \\
Science                             & 0.283 & 0.345 & 0.339 & 0.470 & \textbf{0.240} & 0.264 & 0.273 & 0.633 & 0.684 & 0.258 & 0.258 & 0.348 & \cellcolor[rgb]{0.83,0.89,0.78} 0.366 \\
Self-Help                           & 0.068 & 0.297 & \textbf{0.042} & 0.435 & 0.112 & 0.076 & 0.214 & 0.824 & 0.810 & 0.169 & 0.198 & 0.121 & \cellcolor[rgb]{0.81,0.91,0.78} 0.281 \\
Social Science                      & 0.118 & 0.575 & \textbf{0.057} & 0.384 & 0.370 & 0.252 & 0.065 & 0.661 & 0.540 & 0.128 & 0.300 & 0.222 & \cellcolor[rgb]{0.81,0.90,0.78} 0.306 \\
Sports \& Recreation                & 0.116 & 0.206 & \textbf{0.090} & 0.322 & 0.133 & 0.120 & 0.566 & 0.681 & 0.608 & 0.156 & 0.238 & 0.153 & \cellcolor[rgb]{0.81,0.91,0.78} 0.282 \\
Study Aids                          & \textbf{0.270} & 0.661 & 0.452 & 0.678 & 0.280 & 0.413 & 0.341 & 0.761 & 0.666 & 0.593 & 0.583 & 0.368 & \cellcolor[rgb]{0.87,0.86,0.78} 0.506 \\
Technology \& Engineering           & 0.328 & 0.433 & 0.360 & 0.425 & \textbf{0.294} & 0.330 & 0.506 & 0.723 & 0.752 & 0.409 & 0.397 & 0.320 & \cellcolor[rgb]{0.85,0.87,0.78} 0.440 \\
Transportation                      & \textbf{0.105} & 0.512 & 0.188 & 0.467 & 0.128 & 0.151 & 0.256 & 0.829 & 0.717 & 0.226 & 0.259 & 0.235 & \cellcolor[rgb]{0.82,0.90,0.78} 0.339 \\
Travel                              & 0.467 & 0.493 & \textbf{0.277} & 0.741 & 0.412 & 0.322 & 0.438 & 0.890 & 0.724 & 0.370 & 0.393 & 0.310 & \cellcolor[rgb]{0.86,0.86,0.78} 0.486 \\
True Crime                          & 0.164 & 0.309 & 0.187 & 0.368 & 0.224 & \textbf{0.148} & 0.288 & 0.658 & 0.630 & 0.242 & 0.212 & 0.184 & \cellcolor[rgb]{0.81,0.90,0.78} 0.301 \\
Young Adult Fiction                 & \textbf{0.050} & 0.068 & 0.114 & 0.475 & 0.053 & 0.085 & 0.633 & 0.453 & 0.362 & 0.073 & 0.160 & 0.156 & \cellcolor[rgb]{0.79,0.92,0.78} 0.224 \\
Young Adult Nonfiction              & 0.189 & 0.355 & 0.223 & 0.252 & \textbf{0.168} & 0.187 & 0.281 & 0.655 & 0.747 & 0.203 & 0.195 & 0.235 & \cellcolor[rgb]{0.81,0.90,0.78} 0.307 \\
\bottomrule
\end{tabular}
}
\caption{Text block edit distance by subject (lower is better). \textbf{Bold} = best per row.
         Full model IDs: Claude = claude-opus-4-6; Doubao = doubao-seed-1-6-vision-250815;
         Gemini = gemini-3.1-pro-preview; GPT-4o = gpt-4o; GPT-5.5 = gpt-5.5; Kimi = kimi-k2.5;
         MinerU = mineru2.5; Nemotron = nemotron-nano-12b-v2-vl;
         PaddleOCR = paddleocr-api; Qwen-122B = qwen3.5-122b-a10b;
         Qwen-Flash = qwen3.5-flash; Qwen-Plus = qwen3.5-plus.
         $^\dagger$Music uses a 1-page sliding window; all other subjects use 2 pages.}
\label{tab:text_subject}
\end{table*}
 \begin{figure*}
    \centering
    \includegraphics[width=\linewidth]{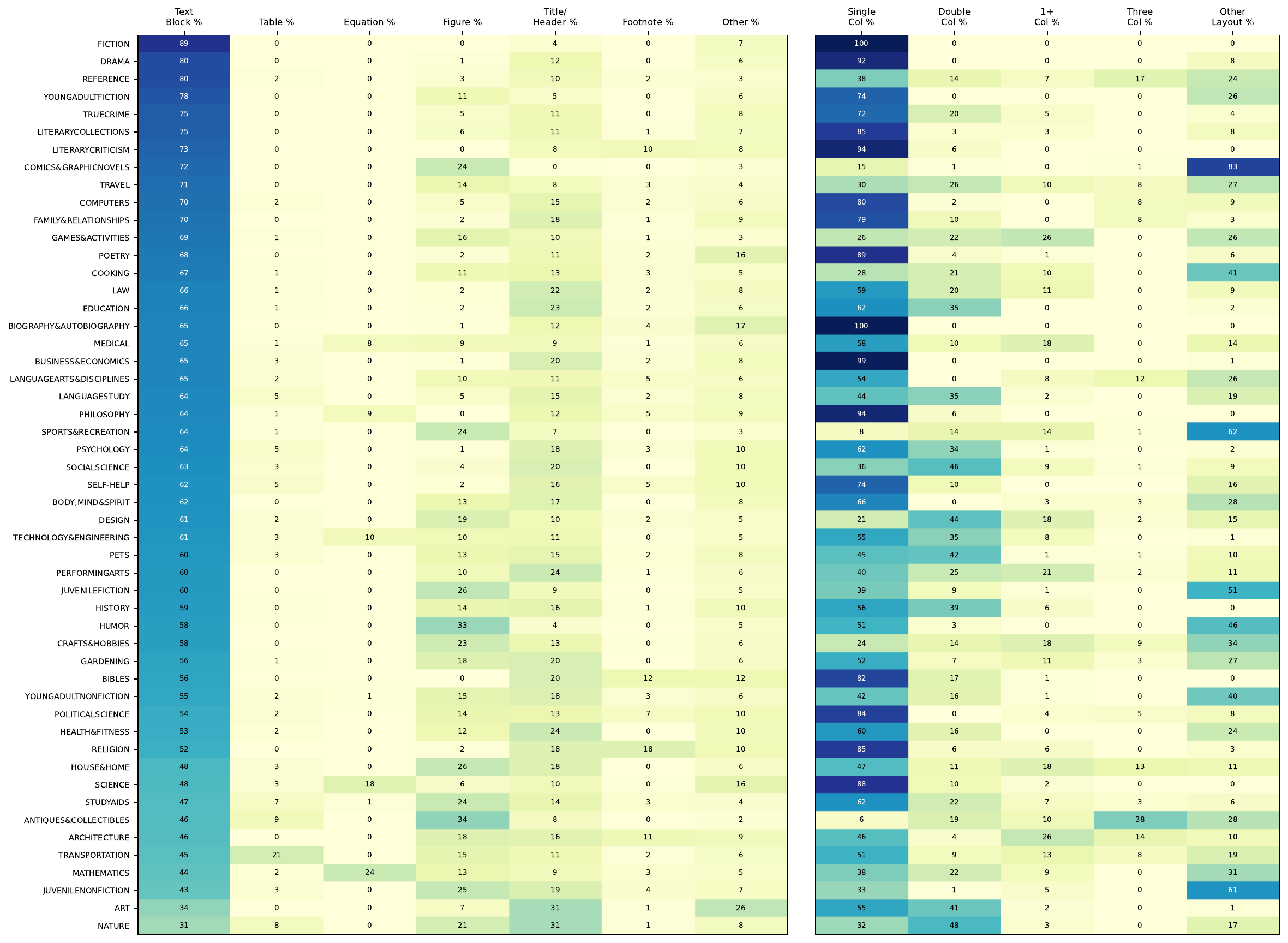}
    \caption{Left: the composition of the data elements per subject. Right: the composition of the layout types per subject. Each individual row sums up to 100\%.}
    \label{fig:per_subj_composition_heatmap}
\end{figure*}
Table~\ref{tab:text_subject} presents the per-subject text edit distance across 52 subject categories. We provide a heatmap of each subject's dominant components and data types in Figure~\ref{fig:per_subj_composition_heatmap}. We separately evaluate Music in Section~\ref{app:music}.

\paragraph{Subject-level wins.}
Kimi achieves the lowest edit distance on 17 of the 51 non-Music subjects, with particular strength on text-heavy genres such as \textsc{Fiction} (0.023), \textsc{Health \& Fitness} (0.039), and \textsc{Psychology} (0.163).
Claude and GPT-5.5 rank second and third with 13 and 11 subject wins, respectively.
GPT-5.5 marks a substantial improvement over GPT-4o: GPT-4o does not lead on any subject and records edit distances 1.5--3$\times$ higher on several categories (e.g., \textsc{Comics \& Graphic Novels}: 0.811 vs.\ 0.295; \textsc{Transportation}: 0.686 vs.\ 0.383).

\paragraph{Model specializations.}
GPT-5.5 leads on subjects with substantial equation content, including \textsc{Mathematics} (0.302), the highest-equation subject (23.9\% of annotated elements), as well as \textsc{Medical} (0.162) and \textsc{Technology \& Engineering} (0.151); Qwen-3.5-122B leads on Science (0.203), the second-highest in equation density (17.5\%).
By contrast, all 13 of Claude's subject wins occur in almost equation-free content (mean equation share: 0.08\%, maximum: 1.0\% in Study Aids), including \textsc{Performing Arts} (0.049), \textsc{Drama} (0.103), \textsc{Transportation} (0.122), \textsc{Literary Criticism} (0.182), and \textsc{Biography \& Autobiography} (0.001).
This correlation suggests Claude is well-calibrated for prose reconstruction but less competitive when mathematical notation is present.

\paragraph{Subject difficulty.}
The lowest-error subjects are text-dense narrative categories: \textsc{Biography \& Autobiography} (Claude: 0.001), \textsc{Business \& Economics} (Kimi: 0.022), and \textsc{Fiction} (Kimi: 0.023).

\subsubsection{Music Analysis}
\label{app:music}
\begin{table}[h!]
  \centering
  \resizebox{\columnwidth}{!}{
  \begin{tabular}{lcc}
  \toprule
  \textbf{Model} & \textbf{Dist} & \textbf{Dist (w/o outlier)} \\
  \midrule
  Baseline     & 0.624& 0.513 \\
  \midrule
  Qwen3.5-flash                  & \textbf{0.662}& \textbf{0.598} \\
  Qwen3.5-122B-A10b              & 0.702 & 0.661 \\
  Kimi-k2.5                      & 0.729 & 0.702 \\
  Qwen3.5-plus                   & 0.762& 0.741 \\
  Gemini         & 0.804 & 0.792 \\
  GPT-5.5                        & 0.901  & 0.895 \\
  Claude                & 0.935 & 0.925 \\
  Doubao  & 0.966 & 0.964 \\
  GPT-4o                         & 0.983 & 0.983 \\
  Nemotron        & 0.993 & 0.994 \\
  Mineru2.5                      & 1.000 & 1.000 \\
  PaddleOCR                & 1.000 & 1.000 \\
  \bottomrule
  \end{tabular}
  }
  \caption{Music score $\rightarrow$ MusicXML transcription results (1-page window). Lower edit distance is better. The baseline is the mean pairwise edit distance across all 6 ground-truth documents. The outlier is the single document with anomalous length (6.1M chars vs.\ 419K–1.3M for the rest).}
  \label{tab:music}
\end{table}

Table~\ref{tab:music} presents the edit distance results for the music score to MusicXML transcription task. This task is not a component of the previous OmniDocBench benchmark, and is uniquely demanding: it requires simultaneous visual reading of musical notation (clefs, key/time signatures, noteheads, articulations, etc.) and structured XML generation that matches the ground-truth schema (see Section~\ref{sec:related} for positioning against prior MLLM music-understanding work). We note up front that our metric is a normalized string-level edit distance over the serialized MusicXML, so it primarily reflects schema-faithful structured generation rather than musical content in isolation; we scope our conclusions accordingly and discuss a content-level metric as a next step below. We treat this subject as a first-of-its-kind exploratory probe over 6 documents rather than a population-level capability estimate.

\paragraph{User-prompt-incapable baselines (minerU2.5, PaddleOCR).} Both score 1 (complete failure) by design. These are pipeline OCR API systems that do not allow a user prompt. Without instruction, they output plain text, Markdown, or JSON, which is maximally distant from MusicXML ground truth. These results are not meaningful as music transcription scores and should be treated as excluded baselines. We still report the numbers for completeness. 

\paragraph{GT cross-document baseline.} To anchor the difficulty of this task, we compute the pairwise edit distance between all 6 groundtruth files (15 pairs), yielding a mean of 0.624 (std 0.194, min 0.378, max 0.932). This is a \emph{null reference} rather than a quality ceiling: it is the score one would obtain by submitting an arbitrary \emph{different} real score, so any model scoring above it produces output no closer to the target than an unrelated GT document would. The high standard deviation (0.194) is driven by extreme length variation across the 6 GT files, which range from 419K to 6.1M characters (a 15$\times$ spread), with one outlier nearly 5$\times$ larger than the next longest, inflating all pairwise distances that involve it. Excluding this outlier yields a mean of 0.513 (std 0.127, 10 pairs), yet the best model score (0.662) still exceeds even this lower null, so the conclusion is robust to the outlier's presence.

\paragraph{Frontier VLMs.} Among the prompt-following models, the Qwen3.5 family dominates the top positions: Qwen3.5-Flash (0.662) achieves the best score. However, this barely clears the null reference (0.624), so even the top model is only marginally closer to the target than a random different GT file. Notably, Qwen3.5-Flash is the smallest but best-performing Qwen variant. Together with a likely ``overthinking penalty'' in larger models, which emit additional commentary, reformatting, or structural variation that inflates string-level edit distance, this points to MusicXML transcription being gated primarily by compact, schema-faithful generation rather than by raw model scale. Gemini 3.1 Pro (0.804) trails the Qwen models but outperforms GPT-5.5 (0.901) and Claude Opus 4.6 (0.935), indicating that the ability to emit schema-faithful MusicXML is unevenly distributed across frontier models. Doubao (0.966) and GPT-4o (0.983) are near-collapsed, approaching the failure threshold of the prompt-incapable baselines.

\paragraph{Effect of the outlier document.} Model rankings are fully preserved with and without the outlier, so all qualitative conclusions stand. However, removing the outlier reveals that the full-corpus scores were understating model failure: the gap between the best model and the null widens from 0.038 (0.662 vs.\ 0.624) to 0.085 (0.598 vs.\ 0.513), because the outlier inflated both model scores and the null mean proportionally — but masked more of the model failure in the process. The Qwen models benefit most from the outlier's removal (Qwen3.5-Flash: $-$0.064, Qwen3.5-122B: $-$0.041), whereas GPT and Claude frontier models drop by $\leq$0.012 and near-failed models (GPT-4o, MinerU2.5, PaddleOCR) are unaffected ($\leq$0.001), indicating they fail uniformly regardless of document length.

\paragraph{Overall takeaway.} On this exploratory probe, no model produces MusicXML closer to the reference than an unrelated score: every system sits at or above the cross-document null under both conditions, so as a schema-faithful transcription task, music remains unsolved for current frontier and specialized systems. Removing the single outlier strengthens this reading, since the best model (Qwen3.5-Flash, 0.598) remains 0.085 above the outlier-excluded null (0.513), a larger margin than the 0.038 observed with all documents. The task cleanly filters out non-prompt-capable systems and separates frontier VLMs along schema-faithful structured generation, a dimension that general document benchmarks do not expose, with the smaller Qwen3.5-Flash leading, suggesting that optimization for structured output generation matters more than scale here. We deliberately scope this probe to schema-faithful generation; isolating musical content from XML verbosity requires a structure-aware score over the parsed MusicXML tree (e.g., TEDS, as we use for tables) or a note-level F1 over (pitch, onset, duration) tuples. We leave this content-level evaluation to future work.

\subsection{The Worst Five Subjects}
\begin{table*}[t]
\centering
\scriptsize
\begin{tabular}{lrrrrrrr}
\toprule
\textbf{Model / Subject} & \textbf{Overall}$\uparrow$ & \textbf{Text}$\downarrow$ & \textbf{Form.}$\downarrow$ & \textbf{CDM}$\uparrow$ & \textbf{TEDS}$\uparrow$ & \textbf{TEDS\textsubscript{S}}$\uparrow$ & \textbf{Order}$\downarrow$ \\
\midrule
\textbf{Claude-Opus-4.6} & 40.33 & 0.469 & 0.414 & 37.86 & 17.66 & 28.54 & 0.473 \\
\cmidrule(l){1-8}
\quad REFERENCE & 20.17 & 0.833 & {--} & {--} & 23.12 & 32.67 & 0.793 \\
\quad DESIGN & 32.06 & 0.634 & {--} & {--} & 15.75 & 18.49 & 0.561 \\
\quad MEDICAL & 47.09 & 0.261 & 0.414 & 37.86 & 18.88 & 33.48 & 0.423 \\
\quad LAW & 50.87 & 0.277 & {--} & {--} & 0.00 & 0.00 & 0.197 \\
\quad GAMES\&ACTIVITIES & 52.47 & 0.341 & {--} & {--} & 30.56 & 58.04 & 0.391 \\
\midrule
\textbf{Doubao} & 29.31 & 0.652 & {--} & {--} & 5.89 & 9.67 & 0.528 \\
\cmidrule(l){1-8}
\quad REFERENCE & 15.69 & 0.909 & {--} & {--} & 4.94 & 11.46 & 0.670 \\
\quad GAMES\&ACTIVITIES & 26.45 & 0.639 & {--} & {--} & 0.84 & 3.24 & 0.576 \\
\quad ANTIQUES\&COLLECTIBLES & 33.30 & 0.593 & {--} & {--} & 3.47 & 5.16 & 0.442 \\
\quad DESIGN & 34.02 & 0.647 & {--} & {--} & 20.21 & 28.48 & 0.534 \\
\quad LAW & 37.09 & 0.470 & {--} & {--} & 0.00 & 0.00 & 0.417 \\
\midrule
\textbf{Gemini-3.1-Pro} & 36.88 & 0.563 & 0.431 & 18.31 & 28.20 & 31.94 & 0.427 \\
\cmidrule(l){1-8}
\quad REFERENCE & 17.83 & 0.860 & {--} & {--} & 25.58 & 33.40 & 0.861 \\
\quad STUDYAIDS & 44.89 & 0.452 & 0.431 & 18.31 & 46.33 & 50.92 & 0.398 \\
\quad LAW & 45.47 & 0.374 & {--} & {--} & 0.00 & 0.00 & 0.262 \\
\quad DESIGN & 48.33 & 0.423 & {--} & {--} & 31.25 & 33.33 & 0.440 \\
\quad LANGUAGEARTS\&DISCIPLINES & 49.98 & 0.704 & {--} & {--} & 37.82 & 42.07 & 0.175 \\
\midrule
\textbf{GPT-4o} & 27.33 & 0.757 & {--} & {--} & 11.51 & 16.78 & 0.538 \\
\cmidrule(l){1-8}
\quad REFERENCE & 25.83 & 0.906 & {--} & {--} & 12.81 & 19.23 & 0.447 \\
\quad GAMES\&ACTIVITIES & 27.73 & 0.654 & {--} & {--} & 3.88 & 8.32 & 0.553 \\
\quad ANTIQUES\&COLLECTIBLES & 27.91 & 0.666 & {--} & {--} & 9.78 & 15.15 & 0.594 \\
\quad COMICS\&GRAPHICNOVELS & 30.53 & 0.866 & {--} & {--} & {--} & {--} & 0.523 \\
\quad DESIGN & 30.97 & 0.694 & {--} & {--} & 19.59 & 24.40 & 0.573 \\
\midrule
\textbf{GPT-5.5} & 44.24 & 0.461 & {--} & {--} & 19.41 & 26.34 & 0.406 \\
\cmidrule(l){1-8}
\quad REFERENCE & 24.98 & 0.788 & {--} & {--} & 27.82 & 37.61 & 0.741 \\
\quad DESIGN & 43.72 & 0.532 & {--} & {--} & 33.28 & 33.33 & 0.489 \\
\quad LAW & 46.01 & 0.383 & {--} & {--} & 0.00 & 0.00 & 0.237 \\
\quad RELIGION & 52.97 & 0.216 & {--} & {--} & 0.00 & 0.00 & 0.195 \\
\quad GAMES\&ACTIVITIES & 53.52 & 0.384 & {--} & {--} & 35.97 & 60.77 & 0.370 \\
\midrule
\textbf{Kimi-K2.5} & 38.19 & 0.550 & 0.367 & 35.76 & 23.73 & 33.36 & 0.517 \\
\cmidrule(l){1-8}
\quad REFERENCE & 25.70 & 0.819 & {--} & {--} & 37.90 & 48.91 & 0.789 \\
\quad DESIGN & 38.29 & 0.620 & {--} & {--} & 35.02 & 39.03 & 0.581 \\
\quad MEDICAL & 42.13 & 0.515 & 0.367 & 35.76 & 30.38 & 45.70 & 0.461 \\
\quad LAW & 42.89 & 0.420 & {--} & {--} & 0.00 & 0.00 & 0.294 \\
\quad GAMES\&ACTIVITIES & 43.90 & 0.375 & {--} & {--} & 15.36 & 33.14 & 0.462 \\
\midrule
\textbf{MinerU2.5} & 30.15 & 0.705 & 0.323 & 31.42 & 25.07 & 32.52 & 0.654 \\
\cmidrule(l){1-8}
\quad COMICS\&GRAPHICNOVELS & 8.39 & 0.962 & {--} & {--} & {--} & {--} & 0.871 \\
\quad REFERENCE & 27.58 & 0.769 & {--} & {--} & 33.63 & 45.73 & 0.740 \\
\quad LAW & 30.62 & 0.649 & {--} & {--} & 0.00 & 0.00 & 0.432 \\
\quad DESIGN & 37.25 & 0.593 & {--} & {--} & 33.14 & 36.79 & 0.621 \\
\quad MEDICAL & 37.29 & 0.552 & 0.323 & 31.42 & 33.52 & 47.58 & 0.606 \\
\midrule
\textbf{Nemotron} & 15.46 & 0.846 & {--} & {--} & 6.45 & 12.73 & 0.755 \\
\cmidrule(l){1-8}
\quad ANTIQUES\&COLLECTIBLES & 12.05 & 0.870 & {--} & {--} & 8.48 & 12.67 & 0.853 \\
\quad CRAFTS\&HOBBIES & 15.37 & 0.814 & {--} & {--} & 2.28 & 15.00 & 0.748 \\
\quad REFERENCE & 15.81 & 0.926 & {--} & {--} & 11.37 & 18.78 & 0.713 \\
\quad LAW & 16.37 & 0.789 & {--} & {--} & 0.00 & 0.00 & 0.720 \\
\quad TRANSPORTATION & 17.69 & 0.829 & {--} & {--} & 10.13 & 17.20 & 0.742 \\
\midrule
\textbf{PaddleOCR} & 14.43 & 0.832 & 0.690 & 8.76 & 16.47 & 29.35 & 0.843 \\
\cmidrule(l){1-8}
\quad DRAMA & 12.13 & 0.871 & {--} & {--} & {--} & {--} & 0.886 \\
\quad REFERENCE & 14.27 & 0.938 & {--} & {--} & 27.47 & 38.96 & 0.909 \\
\quad FAMILY\&RELATIONSHIPS & 14.44 & 0.764 & {--} & {--} & 0.00 & 0.00 & 0.802 \\
\quad COOKING & 15.75 & 0.834 & 1.000 & 0.00 & 26.87 & 44.93 & 0.805 \\
\quad PHILOSOPHY & 18.10 & 0.753 & 0.379 & 17.51 & 11.53 & 33.51 & 0.813 \\
\midrule
\textbf{Qwen3.5-Flash} & 37.19 & 0.591 & 0.312 & 42.07 & 18.05 & 24.95 & 0.523 \\
\cmidrule(l){1-8}
\quad REFERENCE & 24.87 & 0.840 & {--} & {--} & 21.00 & 27.90 & 0.624 \\
\quad LAW & 33.45 & 0.483 & {--} & {--} & 0.00 & 0.00 & 0.513 \\
\quad GAMES\&ACTIVITIES & 37.20 & 0.504 & {--} & {--} & 10.78 & 21.23 & 0.488 \\
\quad DESIGN & 40.17 & 0.603 & {--} & {--} & 32.85 & 35.60 & 0.521 \\
\quad MEDICAL & 42.09 & 0.524 & 0.312 & 42.07 & 25.60 & 40.04 & 0.469 \\
\midrule
\textbf{Qwen3.5-122B} & 34.62 & 0.597 & 0.381 & 22.99 & 22.05 & 28.66 & 0.468 \\
\cmidrule(l){1-8}
\quad STUDYAIDS & 31.14 & 0.593 & 0.420 & 6.40 & 36.86 & 40.77 & 0.594 \\
\quad DESIGN & 33.17 & 0.619 & {--} & {--} & 13.85 & 14.88 & 0.524 \\
\quad GAMES\&ACTIVITIES & 39.36 & 0.465 & {--} & {--} & 8.33 & 17.82 & 0.437 \\
\quad MEDICAL & 39.38 & 0.542 & 0.342 & 39.58 & 20.36 & 30.79 & 0.482 \\
\quad REFERENCE & 41.24 & 0.768 & {--} & {--} & 30.84 & 39.05 & 0.303 \\
\midrule
\textbf{Qwen3.5-Plus} & 36.45 & 0.548 & 0.370 & 38.27 & 15.55 & 24.98 & 0.533 \\
\cmidrule(l){1-8}
\quad REFERENCE & 16.77 & 0.883 & {--} & {--} & 26.81 & 37.52 & 0.882 \\
\quad DESIGN & 32.50 & 0.549 & {--} & {--} & 1.13 & 1.28 & 0.488 \\
\quad LAW & 40.61 & 0.366 & {--} & {--} & 0.00 & 0.00 & 0.416 \\
\quad MEDICAL & 42.94 & 0.506 & 0.370 & 38.27 & 30.10 & 47.90 & 0.460 \\
\quad GAMES\&ACTIVITIES & 44.81 & 0.435 & {--} & {--} & 19.69 & 38.22 & 0.418 \\
\bottomrule
\end{tabular}
\caption{Detailed breakdown per subject results for the worst 5 subjects per model.}
\label{tab:detailed_worst_five}
\end{table*}
\label{app:detailed_worst_five}

The detailed breakdown of the worst five subjects and the individual performance on these five subjects can be found in Table \ref{tab:detailed_worst_five}.

Every single model (all 12) has \textsc{Reference} in its worst 5. This is the one consistent failure across pipeline tools and frontier LLMs, likely due to dense bibliographic entries, unusual list formatting, and high text volume.

\textsc{Design} is also near-universal (10/12 models). Only Nemotron and PaddleOCR escape it, and that's probably because their failure modes are so severe elsewhere that Design doesn't even rank in their worst 5. Heavy visual layout with irregular text placement appears to be a systematic challenge.

Mineru's worst subject is \textsc{Comics}, with an overall score of 8.39. Since Mineru can't receive the user prompt, it never sees our specific prompt for comics. GPT-4o also has Comics in its worst 5, suggesting even with the instruction, it's hard, but the gap between prompted and unprompted models here is enormous.

PaddleOCR's failure mode is categorically different. Its worst subjects are \textsc{Drama}, \textsc{Family \& Relationships}, \textsc{Cooking}, \textsc{Philosophy}, which are more of consumer prose/layout categories rather than the structurally complex subjects like \textsc{Reference} or \textsc{Law} that trip LLMs. Without user prompts. It has no semantic guidance and falls apart on visually rich or narrative-heavy content. Its worst 5 don't overlap much with any other model.

7 out of the 10 LLM-based models have \textsc{Games} in their worst 5. Rulebooks and activity guides likely mix visual layouts, tables, and lists in non-standard ways that confuse even strong models.

\subsection{Additional Analyses}
\label{app:additional_analysis}
\begin{table}[h!]
  \centering
  \small
  \setlength{\tabcolsep}{6pt}
  \begin{tabular}{lrr}
    \toprule
    Model & \makecell{Fuzzy\\Scan$\downarrow$} & \makecell{Color\\BG$\downarrow$} \\
    \midrule
    Claude  & 0.256          & \textbf{0.163} \\
    Doubao  & 0.454          & 0.285          \\
    Gemini   & 0.269          & 0.172          \\
    GPT-4o           & 0.473          & 0.362          \\
    GPT-5.5          & 0.240          & 0.164          \\
    Kimi        & \textbf{0.238} & 0.189          \\
    MinerU       & 0.364          & 0.402          \\
    Nemotron     & 0.665          & 0.573          \\
    PaddleOCR        & 0.774          & 0.655          \\
    Qwen3.5-Flash    & 0.344          & 0.222          \\
    Qwen3.5-122B-A10B & 0.288          & 0.200          \\
    Qwen3.5-Plus      & 0.310          & 0.194          \\
    \bottomrule
  \end{tabular}
  \caption{%
    Text extraction edit distance by page degradation type ($\downarrow$ lower is better). \textbf{Bold}: best per column.}
  \label{tab:text-page-issue}
\end{table}

\paragraph{Page degradation (Table~\ref{tab:text-page-issue}).}
General VLMs consistently score \emph{better} on \texttt{colorful\_background} pages than on \texttt{fuzzy\_scan} pages, suggesting color complexity is less disruptive than scan quality degradation for capable models. MinerU is the only exception: it has a higher edit distance when the background is colorful, making it the only model that degrades more on colored than blurry pages.

\begin{table}
  \centering
  \resizebox{\columnwidth}{!}{%
  \setlength{\tabcolsep}{4pt}
  \begin{tabular}{lrrrrrr}
    \toprule
    Model
      & All
      & \makecell{Single\\Col.}
      & \makecell{Double\\Col.}
      & \makecell{Three\\Col.}
      & \makecell{$\geq$2\\Col.}
      & \makecell{Other\\Layout} \\
    \midrule
    Claude  & 0.188          & 0.158          & 0.233          & 0.202          & 0.256          & 0.241          \\
    Doubao  & 0.283          & 0.213          & 0.409          & 0.454          & 0.384          & 0.326          \\
    Gemini   & 0.210          & 0.177          & 0.255          & 0.293          & 0.269          & 0.258          \\
    GPT-4o           & 0.307          & 0.233          & 0.388          & 0.508          & 0.420          & 0.391          \\
    GPT-5.5          & \textbf{0.167} & \textbf{0.143} & \textbf{0.194} & \textbf{0.180} & \textbf{0.226} & \textbf{0.222} \\
    Kimi        & 0.182          & \textbf{0.133} & 0.231          & 0.274          & 0.263          & 0.271          \\
    MinerU       & 0.300          & 0.237          & 0.340          & 0.336          & 0.357          & 0.463          \\
    Nemotron     & 0.564          & 0.508          & 0.646          & 0.769          & 0.645          & 0.600          \\
    PaddleOCR        & 0.705          & 0.687          & 0.795          & 0.853          & 0.813          & 0.602          \\
    Qwen3.5-Flash    & 0.261          & 0.197          & 0.385          & 0.397          & 0.357          & 0.295          \\
    Qwen3.5-122B     & 0.229          & 0.175          & 0.298          & 0.341          & 0.339          & 0.296          \\
    Qwen3.5-Plus        & 0.260          & 0.222          & 0.336          & 0.365          & 0.329          & 0.280          \\
    \bottomrule
  \end{tabular}
}
\caption{%
    Reading-order edit distance by page layout type ($\downarrow$ lower is better).
    \textbf{Bold}: best per column.}
  \label{tab:order-layout}
\end{table}
\paragraph{Reading order by layout (Table~\ref{tab:order-layout}).}
GPT-5.5 leads on every multi-column layout variant (double 0.194, three 0.180, $\geq$ 0.226, other 0.222), demonstrating strong robustness across complex layouts. Conversely, Doubao degrades sharply as column count increases (single 0.213 $\rightarrow$ double 0.409 $\rightarrow$ three 0.454), suggesting poor generalization to multi-column arrangements. MinerU exhibits the steepest absolute performance decline among non-PaddleOCR models, dropping from single\_column (0.237) to other\_layout (0.463), indicating distinct challenges with non-standard page configurations. However, PaddleOCR's other\_layout score (0.602) surpasses its double\_column (0.795) and three\_column (0.853) performance. %

\begin{table*}
  \centering
  \footnotesize
  \setlength{\tabcolsep}{4pt}
  \begin{tabular}{lrrrrrrrrrrr}
    \toprule
    Model
      & \makecell{Lang\\EN}
      & \makecell{Lang\\ZH}
      & \makecell{Lang\\EN+ZH}
      & \makecell{Lang\\Other}
      & \makecell{BG\\White}
      & \makecell{BG\\Single}
      & \makecell{BG\\Multi}
      & \makecell{Rot.\\Norm.}
      & \makecell{Rot.\\90\textdegree}
      & \makecell{Rot.\\270\textdegree}
      & \makecell{Rot.\\Horiz.} \\
    \midrule
    Claude  & 0.269          & 0.588          & 0.453          & 0.202          & 0.282          & 0.238          & 0.312          & 0.257          & 0.269          & 0.293          & 0.413          \\
    Doubao  & 0.461          & 0.675          & 0.534          & 0.522          & 0.498          & 0.469          & 0.491          & 0.477          & 0.761          & 0.757          & 0.545          \\
    Gemini   & 0.314          & 0.590          & 0.381          & 0.297          & 0.318          & 0.314          & 0.369          & 0.319          & 0.446          & \textbf{0.277} & 0.341          \\
    GPT-4o           & 0.475          & 0.741          & 0.436          & 0.505          & 0.504          & 0.468          & 0.631          & 0.479          & 0.814          & 0.758          & 0.763          \\
    GPT-5.5          & \textbf{0.214} & 0.605          & \textbf{0.168} & \textbf{0.197} & \textbf{0.253} & \textbf{0.191} & \textbf{0.269} & \textbf{0.215} & \textbf{0.240} & 0.419          & \textbf{0.318} \\
    Kimi        & 0.264          & \textbf{0.572} & 0.270          & 0.236          & 0.274          & 0.248          & 0.406          & 0.259          & 0.672          & 0.432          & 0.357          \\
    MinerU       & 0.372          & 0.705          & 0.424          & 0.305          & 0.363          & 0.345          & 0.637          & 0.354          & 0.539          & 0.563          & 0.546          \\
    Nemotron     & 0.671          & 0.877          & 0.537          & 0.813          & 0.727          & 0.698          & 0.764          & 0.708          & 0.880          & 0.750          & 0.964          \\
    PaddleOCR        & 0.938          & 0.899          & 0.947          & 0.945          & 0.935          & 0.942          & 0.930          & 0.941          & 0.971          & 0.966          & 0.854          \\
    Qwen3.5-Flash    & 0.402          & 0.714          & 0.499          & 0.404          & 0.411          & 0.412          & 0.387          & 0.408          & 0.518          & 0.576          & 0.478          \\
    Qwen3.5-122B     & 0.321          & 0.654          & 0.328          & 0.299          & 0.344          & 0.302          & 0.398          & 0.319          & 0.496          & 0.481          & 0.432          \\
    Qwen3.5-Plus        & 0.393          & 0.675          & 0.430          & 0.448          & 0.437          & 0.398          & 0.389          & 0.416          & 0.502          & 0.527          & 0.370          \\
    \bottomrule
  \end{tabular}
  \caption{%
    Text extraction edit distance by text-block element attributes ($\downarrow$ lower is better).
    \emph{Lang}: language label of the individual text block.
    \emph{BG}: background color (White / Single-colored / Multi-colored).
    \emph{Rot.}: text rotation (Normal / 90\textdegree{} / 270\textdegree{} / Horizontally flipped).
    \textbf{Bold}: best per column.}
  \label{tab:text-element}
\end{table*}

\paragraph{Text element attributes by edit distance (Table~\ref{tab:text-element}).}
Rotation is a sharp model differentiator. GPT-4o is the most rotation-sensitive frontier VLM, with rotate90 (0.814) and rotate270 (0.758) far exceeding its normal score (0.479). Doubao similarly collapses on rotate90 (0.761) and rotate27z0 (0.757). Kimi shows an asymmetric pattern: rotate90 (0.672) is catastrophic while rotate270 (0.432) is tolerated, suggesting partial handling. GPT-5.5 is the most rotation-robust frontier model (0.240 / 0.419). Chinese-script text elements (\texttt{text\_simplified\_chinese}) are substantially harder than English elements across all models: kimi's ZH score (0.572) is 2.2$\times$ its EN score (0.264), and claude's ZH (0.588) is 2.2$\times$ its EN (0.269). This is in apparent tension with Table~\ref{tab:text-lang}, where Claude and Kimi score far better on Chinese-language \emph{pages}; the discrepancy arises because the two breakdowns measure different slices --- element-level language vs.\ page-level document language --- and is discussed further in the context of Table~\ref{tab:text-lang}. MinerU shows the largest degradation under multi-colored text backgrounds (0.637 vs.\ 0.363 for white), steeper than any frontier VLM.

\begin{table*}
  \centering
  \small
  \setlength{\tabcolsep}{5pt}
  \begin{tabular}{lrrrrrrrr}
    \toprule
    Model
      & All
      & \makecell{Acad.\\Lit.}
      & \makecell{Color.\\TB}
      & \makecell{Exam\\Paper}
      & Book
      & \makecell{Single\\Col.}
      & \makecell{Double\\Col.}
      & \makecell{Other\\Layout} \\
    \midrule
    Claude  & 0.368          & 0.253          & 0.891          & 0.888          & \textbf{0.259} & 0.273          & 0.318          & 0.940          \\
    Doubao  & 0.313          & 0.208          & 0.719          & 0.527          & 0.277          & 0.279          & \textbf{0.201} & 0.545          \\
    Gemini   & 0.353          & 0.249          & 0.724          & 0.757          & 0.280          & 0.290          & 0.277          & 0.775          \\
    GPT-4o           & 0.468          & 0.381          & 0.906          & 0.641          & 0.440          & 0.413          & 0.645          & 0.708          \\
    GPT-5.5          & 0.359          & 0.250          & 0.449          & 0.803          & 0.289          & 0.294          & 0.263          & 0.860          \\
    Kimi       & 0.350          & 0.177          & 0.741          & 0.769          & 0.277          & 0.274          & 0.247          & 0.842          \\
    MinerU       & 0.506          & 0.470          & 0.518          & 0.885          & 0.439          & 0.454          & 0.288          & 0.952          \\
    Nemotron     & 0.759          & 0.604          & 0.844          & 0.955          & 0.751          & 0.724          & 0.703          & 0.969          \\
    PaddleOCR        & 0.415          & 0.395          & \textbf{0.315} & \textbf{0.442} & 0.417          & 0.412          & 0.461          & \textbf{0.398} \\
    Qwen3.5-Flash    & 0.317          & 0.179          & 0.773          & 0.678          & 0.262          & \textbf{0.260} & 0.232          & 0.697          \\
    Qwen3.5-122B     & 0.322          & \textbf{0.173} & 0.348          & 0.574          & 0.282          & 0.286          & 0.217          & 0.576          \\
    Qwen3.5-Plus         & \textbf{0.307} & 0.190          & 0.660          & 0.579          & 0.265          & 0.261          & 0.264          & 0.600          \\
    \bottomrule
  \end{tabular}
  \caption{%
    Display formula edit distance broken down by data source and page layout
    ($\downarrow$ lower is better).
    \emph{Color.~TB}: colorful textbook;
    \emph{Acad.~Lit.}: academic literature.
    \textbf{Bold}: best per column.}
  \label{tab:formula-breakdown}
\end{table*}

\paragraph{Formula performance by document type and layout (Table~\ref{tab:formula-breakdown}).}
GPT-4o's formula recognition degrades most severely on \texttt{colorful\_textbook} (0.906), followed closely by Claude (0.891) and \texttt{exam\_paper} (0.888) --- a 3.5$\times$ decline from Claude's \texttt{academic\_literature} score (0.253). Claude's other\_layout formula edit distance (0.940) is the worst among frontier VLMs (GPT-5.5: 0.860, kimi: 0.842, gemini: 0.775), indicating a particular weakness with formulas in non-standard page arrangements. PaddleOCR leads on \texttt{colorful\_textbook} (0.315) and \texttt{other\_layout} (0.398) despite its \texttt{\$\$} delimiter penalty (Section~\ref{sec:experiments}), further confirming that its formula recognition module is genuinely strong on these conditions.

\begin{table}[t] 
  \centering
  \small
  \setlength{\tabcolsep}{4pt}
  \resizebox{\columnwidth}{!}{%
  \begin{tabular}{lrrrr}
    \toprule
    Model
      & All
      & \makecell{Percep-\\tion}
      & \makecell{Structural\\Recon.}
      & \makecell{Domain\\Reasoning} \\
    \midrule
    Claude  & 0.217          & 0.158          & 0.256          & \textbf{0.123} \\
    Doubao  & 0.344          & 0.270          & 0.392          & 0.402          \\
    Gemini   & 0.219          & 0.167          & 0.255          & 0.192          \\
    GPT-4o           & 0.366          & 0.293          & 0.414          & 0.284          \\
    GPT-5.5          & \textbf{0.189} & 0.148          & \textbf{0.223} & 0.146          \\
    Kimi     & 0.193          & \textbf{0.137} & 0.234          & 0.165          \\
    MinerU     & 0.325          & 0.249          & 0.387          & 0.236          \\
    Nemotron    & 0.622          & 0.570          & 0.656          & 0.644          \\
    PaddleOCR        & 0.729          & 0.700          & 0.738          & 0.668          \\
    Qwen3.5-Flash    & 0.258          & 0.190          & 0.303          & 0.266          \\
    Qwen3.5-122B     & 0.226          & 0.163          & 0.271          & 0.292          \\
    Qwen3.5-Plus        & 0.249          & 0.203          & 0.280          & 0.285          \\
    \bottomrule
  \end{tabular}}
  \caption{%
    Text extraction edit distance by challenge type ($\downarrow$ lower is better).
    \textbf{Bold}: best per column.}
  \label{tab:text-challenge}
\end{table}

\paragraph{Challenge type (Table~\ref{tab:text-challenge}).}
Most models score better on \texttt{domain\_reasoning} than on \texttt{structural\_reconstruction} despite the former being semantically harder by design --- including PaddleOCR (domain 0.668 vs.\ structural 0.738), which has no language understanding. This suggests the category labels partly reflect the visual and structural complexity of individual elements: domain\_reasoning elements may be shorter or less layout-intensive, making them easier to transcribe accurately regardless of semantic content. Doubao, Qwen3.5+, and Qwen3.5-122B are exceptions that show domain $>$ structural (worse performance on the semantically harder type), contrasting with claude and GPT-5.5, where domain\_reasoning is the \emph{best}-scoring challenge type.

\begin{table}[t] 
  \centering
  \resizebox{\columnwidth}{!}{%
  \setlength{\tabcolsep}{5pt}
  \begin{tabular}{lrrrr}
    \toprule
    Model & All & English & \makecell{Simplified\\Chinese} & Other \\
    \midrule
    Claude & 0.217          & 0.218          & \textbf{0.053} & 0.211          \\
    Doubao & 0.344          & 0.293          & 0.353          & 0.521          \\
    Gemini   & 0.219          & 0.200          & 0.055          & 0.264          \\
    GPT-4o           & 0.366          & 0.321          & 0.541          & 0.493          \\
    GPT-5.5          & \textbf{0.189} & \textbf{0.156} & 0.179          & \textbf{0.204} \\
    Kimi       & 0.193          & 0.173          & 0.059          & 0.205          \\
    MinerU      & 0.325          & 0.293          & 0.393          & 0.340          \\
    Nemotron     & 0.622          & 0.531          & 0.752          & 0.832          \\
    PaddleOCR        & 0.729          & 0.732          & 0.628          & 0.732          \\
    Qwen3.5-Flash    & 0.258          & 0.241          & 0.288          & 0.307          \\
    Qwen3.5-122B     & 0.226          & 0.210          & 0.210          & 0.270          \\
    Qwen3.5-Plus        & 0.249          & 0.222          & 0.248          & 0.300          \\
    \bottomrule
  \end{tabular}
  }
  \caption{%
    Text extraction edit distance by page-level language ($\downarrow$ lower is better).
    \textbf{Bold}: best per column.}
  \label{tab:text-lang}
\end{table}

\paragraph{Page-level language (Table~\ref{tab:text-lang}).}
Models are split into three groups based on Chinese-page performance. Claude, Kimi, and Gemini score dramatically lower in edit distance (better) on Chinese pages than English (Claude: 0.218 $\to$ 0.053; Kimi: 0.173 $\to$ 0.059; Gemini: 0.200 $\to$ 0.055 --- approximately 4$\times$ improvement). GPT-5.5 and most other models show the opposite or no advantage: GPT-5.5 is slightly worse on Chinese (0.156 $\to$ 0.179), as are doubao and mineru. Separately, the element-level breakdown (Table~\ref{tab:text-element}) shows that Chinese-script text elements are substantially \emph{harder} for all models --- Kimi's ZH element score (0.572) is 2.2$\times$ its EN score (0.264). These two observations are not directly comparable: Table~\ref{tab:text-lang} aggregates all content on a page by that page's primary language, while Table~\ref{tab:text-element} groups individual blocks by their own script label. The underlying causes of the page-level advantage for the first model group and the element-level hardness of Chinese-script text are open questions that require inspecting the document-type distribution of Chinese-language pages to resolve.

\subsection{Scaling of model size}
\begin{figure*}[ht!]
    \centering
    \includegraphics[width=\linewidth]{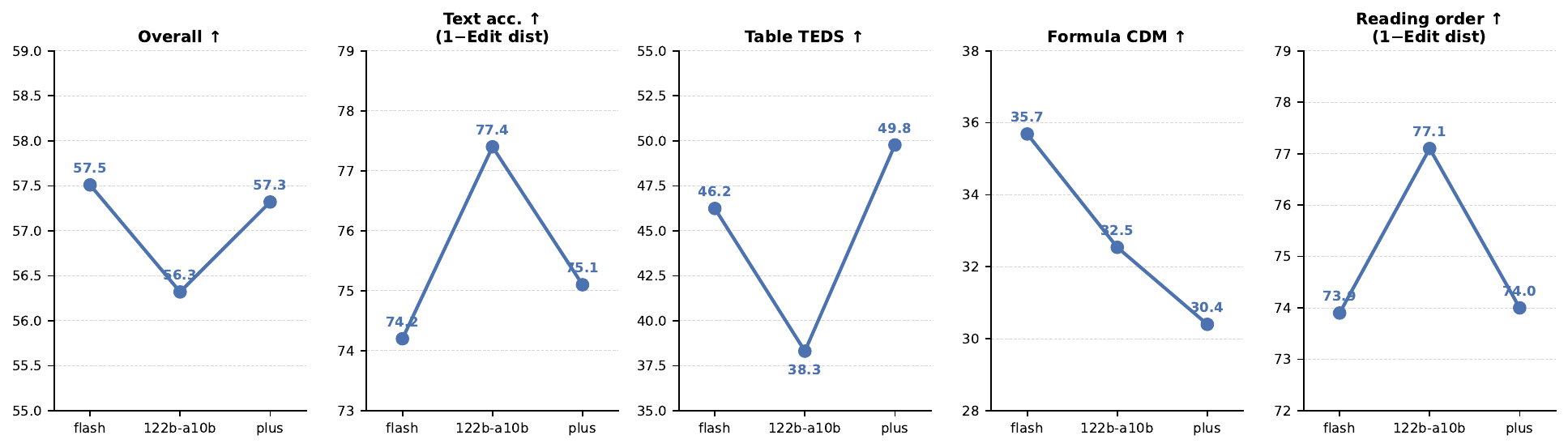}
    \caption{Scaling per model size}
    \label{fig:scaling_size}
\end{figure*}

We analyze the impact of model size within the Qwen3.5 family, comparing three variants ordered by scale: flash (smallest), 122B-A10B, and plus (largest). Figure~\ref{fig:scaling_size} reports performance across five metrics. The overall scores vary within a narrow range (57.5, 56.3, and 57.3 for the three variants, respectively), but individual metrics exhibit larger variation.

\paragraph{Formula CDM decreases monotonically with scale.} Formula recognition accuracy decreases as model size increases: 35.7 (flash) $\to$ 32.5 (122B-A10B) $\to$ 30.4 (plus), a cumulative drop of 5.3 points. This trend suggests a trade-off between formula-level precision and other capabilities.

\paragraph{Per-metric trade-offs across model variants.} Table TEDS drops from 46.2 (flash) to 38.3 (122B-A10B), a decrease of 7.9 points, before recovering to 49.8 (plus). Conversely, 122B-A10B achieves the highest text accuracy (77.4 vs.\ 74.2 for flash) and reading-order accuracy (77.1 vs.\ 73.9 for flash). These inverse rankings indicate that the mid-tier model performs better on text transcription and ordering than on structured layout recognition.

These results indicate that scaling model size does not uniformly improve document understanding; per-metric trends can move in opposite directions even as the overall score remains stable. We note a potential confound: Qwen3.5-Flash and Qwen3.5-Plus are production API variants with possible serving-side enhancements and extended context window,~\footnote{Qwen3.5-Plus: \url{https://qwen.ai/blog?id=qwen3.5}, Qwen3.5-Flash: \url{https://huggingface.co/Qwen/Qwen3.5-35B-A3B}. Both with a 1M context length.} whereas the 122B-A10B variant (32k context length) is served without such optimizations, which may partly account for the observed differences.

\subsection{Scaling of sliding window size}
\label{app:sliding_window}
\begin{figure}
    \centering
    \includegraphics[width=\linewidth]{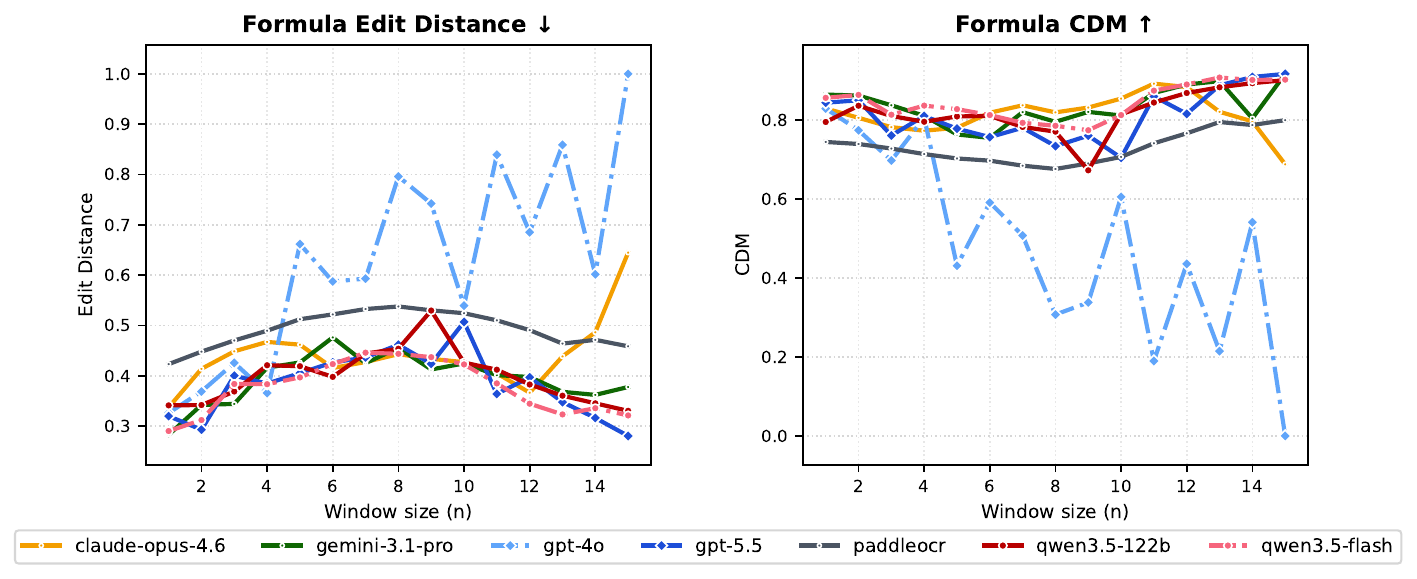}
    \caption{Impact of sliding window size $n$ pages) on the formula metrics across seven models.}
    \label{fig:formula_sliding_window}
\end{figure}
We evaluate how varying the sliding window size $n$ (from 1 to 15 pages) affects nine models: Claude, Gemini, GPT-4o, GPT-5.5, Qwen-3.5 flash (smallest), 122B-A10B, and plus (largest), as well as two specialized VLMs (MinerU2.5 and PaddleOCR). Based on the subject hardness, we arbitrarily choose to focus on two subjects: law and medicine.
Figure~\ref{fig:scaling_window} shows the aggregate score and per-metric trends, Figure~\ref{fig:formula_sliding_window} shows the formula metrics trends.

\paragraph{Reading order degrades most consistently.} The most robust correlation across all models and subjects is between window size and reading order edit distance. Aggregated over all subjects, Claude's reading order error rises from 0.24 at $n{=}1$ to 0.69 at $n{=}15$; GPT-5.5 climbs from 0.17 to 0.47; Gemini from 0.24 to 0.51. GPT-4o shows the sharpest degradation among frontier LLMs, rising from 0.38 to 0.85. The effect is especially pronounced in the law subject, where Claude's reading order edit distance increases from 0.077 at $n{=}1$ to 0.694 at $n{=}15$. Larger windows require the model to track ordering cues across more pages, and accuracy tends to fall as this span grows.

\paragraph{Text edit distance shows a mixed early-window benefit.} For text edit distance, GPT-5.5 and Gemini both improve marginally from $n{=}1$ to $n{=}2$ (GPT-5.5: $0.287 \to 0.285$; Gemini: $0.335 \to 0.322$), suggesting a slight cross-page context benefit for resolving boundary-spanning content. Claude generally worsens with context size ($0.220$ at $n{=}1$ to $0.365$ at $n{=}15$), while GPT-4o shows the most severe degradation among frontier models, rising from $0.531$ to $0.782$ at $n{=}15$. Qwen-Plus shows relatively flat behavior throughout ($0.345$ at $n{=}1$ to $0.369$ at $n{=}15$). Specialized VLMs (MinerU2.5 and PaddleOCR) are nearly invariant across all window sizes, consistent with pipeline architectures that do not leverage cross-page context. Beyond $n{=}2$, text accuracy generally worsens across all models, consistent with longer contexts introducing more opportunities for content confusion or truncation.

\paragraph{Formula metrics show no systematic degradation.} Formula CDM stays within a moderate band for most models across small-to-moderate window sizes (e.g., Qwen-Plus: $0.817$ at $n{=}1$ to a range of $0.763$–$0.867$ through $n{=}12$). Claude shows a substantial drop at very large windows ($0.687$ at $n{=}15$). GPT-4o begins failing erratically from $n{=}5$ onward—oscillating between near-zero values and partial recoveries before collapsing completely to $0.000$ at $n{=}15$. In stark contrast, GPT-5.5 remains broadly stable throughout ($0.704$–$0.916$), suggesting a significant difference in how the two models handle structured element prediction under long contexts.

\paragraph{The $n{=}2$ default.} The two-page window used throughout our main evaluation reflects a practical trade-off. It captures enough cross-page context to benefit models that improve from $n{=}1$ to $n{=}2$, while staying well below the window sizes where reading order and text accuracy degrade substantially. Larger windows yield diminishing returns at best and measurably worse performance at realistic document lengths.

\end{document}